%% file: main.tex
\documentclass[letterpaper,twocolumn,10pt]{article}
\usepackage{usenix2019_v3}
\input{settings}

\begin{document}

\date{}

\title{\Large \bf \textit{AttackGNN}:\\
Red-Teaming GNNs in Hardware Security Using Reinforcement Learning}

\author{
{\rm Vasudev Gohil$^{\dagger,}$, Satwik Patnaik$^{\ddagger}$, Dileep Kalathil$^\dagger$, Jeyavijayan (JV) Rajendran$^\dagger$}\\
$^\dagger$Texas A\&M University, USA, $^\ddagger$University of Delaware, USA\\
{\tt $^\dagger$\{gohil.vasudev, dileep.kalathil, jv.rajendran\}@tamu.edu,} {\tt $^\ddagger$satwik@udel.edu}
} 

\maketitle

\begin{abstract}
\input{abstract}

\end{abstract}

\input{texfiles/introduction}

\input{texfiles/background}

\input{texfiles/threat_model}
\input{texfiles/methodology}
\input{texfiles/results}

\input{texfiles/related_work_and_discussion}
\input{texfiles/conclusion}
\input{texfiles/acknowledgement}

\bibliographystyle{plainone}
\bibliography{main}

\input{texfiles/appendix}

\end{document}

%% file: settings.tex
\usepackage{xcolor}

\usepackage{soul}
\newcommand{\drop}[1]{\textcolor{red}{#1}}
\renewcommand{\drop}[1]{}
\definecolor{cadmiumgreen}{rgb}{0.0, 0.42, 0.24}
\usepackage{multirow}
\usepackage{enumitem}

\usepackage{graphicx}
\usepackage{booktabs} 
\usepackage{pifont}
\usepackage{mathtools}
\def\smallerspacecaption{\vspace{-2mm}}

\usepackage[ruled, vlined, norelsize]{algorithm2e} 
\SetKwInput{KwInput}{Input}
\SetKwInput{KwOutput}{Output}
\SetKwInput{KwInit}{Initialization}
\SetKwInput{Kwprocedure}{Procedure}
\SetKwInput{KwParam}{Parameter}

\usepackage{lipsum}
\usepackage{dblfloatfix}

\usepackage{bm}

\usepackage{wasysym}

\newcommand{\myname}{\textit{AttackGNN}}

\usepackage{hyperref}
\usepackage{breakurl}

\usepackage{tikz}
\newcommand*\cc[1]{\tikz[baseline=(char.base)]{
\textcolor{white}{      \node[shape=circle,draw,inner sep=1pt,fill=black] (char) {#1};}}}

\usepackage[framemethod=default]{mdframed}
\newmdenv[linewidth=1pt, linecolor=black,backgroundcolor=gray!15]{myframe}

\definecolor{codecommentcolor}{RGB}{0,124,0}

\usepackage{caption}

%% file: abstract.tex
Machine learning has shown great promise in addressing several critical hardware security problems. In particular, researchers have developed novel graph neural network (GNN)-based techniques for detecting intellectual property (IP) piracy, detecting hardware Trojans (HTs), and reverse engineering circuits, to name a few. These techniques have demonstrated outstanding accuracy and have received much attention in the community. 
However, since these techniques are used for security applications, it is imperative to evaluate them thoroughly and ensure they are robust and do not compromise the security of integrated circuits. 

In this work, we propose~\myname, the first red-team attack on GNN-based techniques in hardware security. To this end, we devise a novel reinforcement learning (RL) agent that generates adversarial examples, i.e., circuits, against the GNN-based techniques. We overcome three challenges related to effectiveness, scalability, and generality to devise a potent RL agent.
We target five GNN-based techniques for four crucial classes of problems in hardware security: IP piracy, detecting/localizing HTs, reverse engineering, and hardware obfuscation. 
Through our approach, we craft circuits that fool all GNNs considered in this work.
For instance, to evade IP piracy detection, we generate adversarial pirated circuits that fool the GNN-based defense into classifying our crafted circuits as not pirated. For attacking HT localization GNN, our attack generates HT-infested circuits that fool the defense on all tested circuits. We obtain a similar $100\%$ success rate against GNNs for all classes of problems.

%% file: texfiles/introduction.tex
\section{Introduction}\label{sec:introduction}

\subsection{Threats Due to Globalized IC Supply Chain}
\label{sec:globalized_supply_chain}

Modern computing systems heavily rely on integrated circuits (ICs), which serve as their foundation. To achieve high performance and low power consumption in ICs, it is essential to have access to smaller and faster transistors, which are the basic components of ICs. The ongoing drive to continuously shrink transistors necessitates using cutting-edge fabrication facilities, commonly known as foundries. However, the cost of employing such advanced foundries is exorbitant. 
For instance, Samsung recently announced that it plans to invest $\$228$ billion in a new semiconductor complex in South Korea, which will be the world’s largest~\cite{samsung_228_billion}.
To address the challenges of design costs and overcome the tight time-to-market constraints, prominent IC design companies like NVIDIA and Apple
operate under a fabless model. They outsource IC manufacturing to offshore third-party foundries, introducing potential trust concerns. 
In the U.S. Department of Defense's strategy for safeguarding critical defense supply chains in 2022, it was disclosed that a substantial $88\%$ of microelectronic manufacturing takes place outside the U.S., thereby presenting a notable security concern~\cite{DoD_supply_chain}.
This distributed supply chain arrangement has resulted in numerous security issues, including intellectual property (IP) piracy~\cite{alkabani2007active,imeson2013securing} and the insertion of malicious logic called hardware Trojans (HTs)~\cite{agrawal2007trojan,rostami2014primer,hicks2010overcoming,waksman2013fanci,yang2016a2,trippel2021bomberman,gohil2022attrition}.

\subsection{Impact of Hardware Security Problems}

Hardware security problems such as IP piracy, HTs, and reverse engineering profoundly impact various aspects of technology and security. 
For instance, in 2018, as the U.S. Department of Justice reported, the global market for dynamic random-access memory (DRAM) was valued at nearly \$100 billion. Micron, a major player in the DRAM industry holding a 20-25\% market share, incurred an estimated loss of \$8.75 billion due to IP piracy, underscoring the significant economic impact of IP piracy~\cite{Mircon_IP_piracy}. HTs are another serious threat to the security of ICs.
HTs can cause denial-of-service, privilege escalation, or leak confidential information. For instance, researchers discovered a ``backdoor'' in a military-grade chip~\cite{skorobogatov2012breakthrough}. 
Researchers have also demonstrated HTs that can compromise the security of Intel's Ivy Bridge processors~\cite{becker2013stealthy} or cause privilege escalation using capacitor-based HTs on fabricated chips~\cite{yang2016a2}. The examples shown above underscore the pernicious consequences of IP piracy and HTs, prompting research efforts by organizations like the Defense Advanced Research Projects Agency (DARPA) to counteract these threats using programs such as the Automatic Implementation of Secure Silicon program~\cite{aiss}.

\subsection{Graph Neural Networks in Hardware Security}\label{sec:motivation_GNNs_in_hw_sec}

As explained above, industry players, such as Intel, Qualcomm, Synopsys, Cadence, etc., and government agencies, such as DARPA, are investing a lot of effort into not only the power, performance, and area aspects of computing systems but also the security of those systems~\cite{companies_pufs,synopsys_pufs,cadence_pufs}. To aid this process of securing hardware, researchers have recently utilized graph neural networks (GNNs) for several hardware security-related tasks, showcasing state-of-the-art performance in identifying IP piracy~\cite{GNN4IP}, detecting and locating HTs~\cite{GNN4TJ,TrojanSAINT_ISCAS}, reverse engineering circuits~\cite{GNNRE}, and breaking hardware obfuscation techniques~\cite{OMLA,alrahis2022muxlink}, among others. However, there exists a crucial gap in using such GNN-based techniques for hardware security: these techniques have not been evaluated thoroughly. In particular, the threat of adversarial attacks on ML-based systems is extremely pernicious and must be understood and mitigated effectively. For instance, if a GNN that has not been thoroughly evaluated for adversarial robustness is used in detecting IP piracy, it can incorrectly classify a pirated circuit as not pirated, which can lead to a tremendous loss for the IP design house. Similarly, if a GNN that has not been thoroughly evaluated for adversarial robustness is used in detecting HTs, it can incorrectly classify an HT-infested circuit as HT-free, which can lead to disastrous consequences such as compromised encryption security.

\begin{table*}[t]
\centering
\caption{\myname{} against GNNs used in hardware security. 
}
\smallerspacecaption
\resizebox{\textwidth}{!}{
\begin{tabular}{cccccc}
\toprule
\textbf{Technique Type} & \multicolumn{3}{c}{Defense} & \multicolumn{2}{c}{Attack} \\
\cmidrule(r){1-1} \cmidrule(lr){2-4} \cmidrule(l){5-6}
\textbf{Security Problem} & Detecting HTs & Localizing HTs & Detecting IP Piracy & Reverse Engineering & Hardware Obfuscation\\
\cmidrule{1-6}
\textbf{Technique} & GNN4TJ~\cite{GNN4TJ} & TrojanSAINT~\cite{TrojanSAINT_ISCAS} & GNN4IP~\cite{GNN4IP} & GNN-RE~\cite{GNNRE} & OMLA~\cite{OMLA} \\ \cmidrule{1-6}
\textbf{GNN Framework} & \begin{tabular}[c]{@{}c@{}}Attention-based\\ custom GCN\end{tabular} & \begin{tabular}[c]{@{}c@{}}Graph attention network~\cite{GAT}\\ (w. GraphSAINT~\cite{GraphSAINT} for sampling)\end{tabular} & \begin{tabular}[c]{@{}c@{}}Attention-based\\custom GCN\end{tabular} & \begin{tabular}[c]{@{}c@{}}Graph attention network~\cite{GAT}\\ (w. GraphSAINT~\cite{GraphSAINT} for sampling)\end{tabular} & \begin{tabular}[c]{@{}c@{}}Graph isomorphism\\network~\cite{GIN}\end{tabular} \\ \cmidrule{1-6}
\textbf{Claimed Efficacy} & $97\%$ TPR & $98\%$ TPR, $96\%$ TNR & $94.61\%$ Acc. & $98.87\%$ Acc. & $89.55\%$ Acc.\\ \cmidrule{1-6}
\textbf{\begin{tabular}[c]{@{}c@{}}\myname~(This Work)'s\\ Adversarial Success Rate\end{tabular}} & $100$\% & $100$\% & $100$\% & $100$\% & $100$\% \\
\bottomrule
\end{tabular}
}
\label{tab:overview_table}
\smallerspacecaption
\smallerspacecaption
\end{table*}

\subsection{Our Contributions}\label{sec:contributions}

In this work, we address the above-mentioned research gap using \myname{}, which performs a thorough evaluation of the GNN-based techniques in hardware security. To do so, we devise adversarial examples, i.e., circuits, against GNNs in hardware security for problems ranging from (i) detecting IP Piracy, (ii) detecting/localizing HTs, (iii) reverse engineering circuits, to (iv) breaking hardware obfuscation techniques for protecting circuit functionality.
However, the threat model of devising adversarial examples places strict constraints (e.g., black-box access) on the attacker.
The challenges for adversarial example generation are further exacerbated due to our field of application of GNNs, i.e., hardware security. Since we work with circuits, which need to obey design rule constraints, unlike arbitrary graphs, traditional perturbation-based adversarial example techniques, such as adding/deleting edges, injecting nodes, or modifying features, are not suitable in our case.
Additionally, typical circuits consist of several thousands of gates, i.e., nodes, and even more wires, i.e., edges. Such a large design space of circuits makes the problem even more challenging. Simply brute-forcing all combinations of perturbations is clearly impossible. For instance, if we just restrict to perturbations that delete two edges in a graph with $1000$ edges (a small circuit), the possible combinations are $^{1000}C_2 = 499,500$. Another practical consideration required when working with such large circuits is that performing operations on them is expensive. For instance, synthesizing or resynthesizing (i.e., compiling) large circuits can take several minutes. Thus, to ensure a practical technique, a balance needs to be struck in the trade-off between runtime and efficacy for such circuits. Likewise, testing adversarially-perturbed large circuits is also expensive since GNN-based tools require more time to analyze them. An appropriate trade-off must also be made in this regard (more details about this are provided in Sec.~\ref{sec:challenge_and_solution_2}). Moreover, different circuits have vastly different structures. For instance, an encryption circuit will have very different gates and connections between them (convoluted operations for ensuring encryption security) compared to an adder circuit. This means that perturbations that work for one circuit may perform poorly for other circuits. These hurdles (large design space exploration and difficulty in generalizing to various circuits) make it challenging to devise successful adversarial examples.

\begin{figure}
    \centering
    \includegraphics[width=0.48\textwidth,trim={1cm 0.5cm 0.7cm 0.5cm},clip]{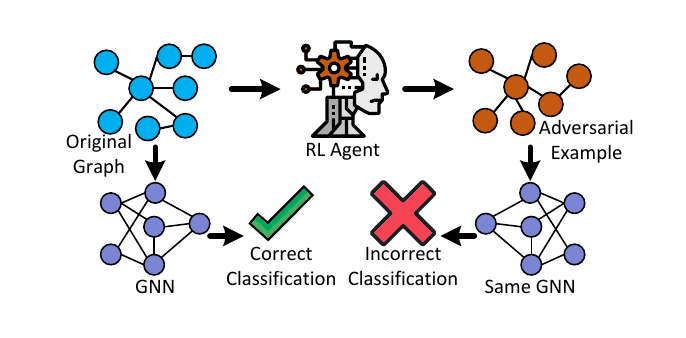}
    \smallerspacecaption
    \smallerspacecaption
    \smallerspacecaption
    \caption{High-level overview of the proposed RL-based adversarial example attack against GNNs 
    in hardware security.}
    \label{fig:high_level_idea}
    \smallerspacecaption
    \smallerspacecaption
\end{figure}

We address these hurdles by modeling the adversarial example generation problem as a Markov decision process (MDP) and solving it using reinforcement learning (RL). 
RL has shown great promise in large design-space exploration by navigating unknown and uncertain problem spaces and finding optimal or near-optimal solutions.
However, a straightforward application of RL is not sufficient to generate high-quality adversarial examples. Hence, we investigate and optimize the RL agent on three fronts: \cc{1} designing effective and generalizable actions, i.e., functionality preserving perturbations to the circuits, \cc{2} sparse rewards for faster training, i.e., ability to scale to larger circuits, and \cc{3} enabling multi-task learning using contextual MDPs, i.e., a single RL agent to generate successful adversarial examples against all GNNs. Incorporating all three optimizations results in an automated, generic, and practical method that evaluates the susceptibility of GNN-based techniques in hardware security to adversarial examples (details in Sec.~\ref{sec:methodology}). 
The primary contributions of this work are:
\begin{itemize}[leftmargin=*]
    \item We develop a first-of-its-kind RL-based adversarial example generation technique,~\myname,\footnote{To enable future research on the practicality of GNNs in hardware security, we have open-sourced our adversarial examples to the community \href{https://github.com/gohil-vasudev/AttackGNN}{here}.} for GNNs used in hardware security. Figure~\ref{fig:high_level_idea} illustrates the high-level concept of this work.
    \item \myname~is agnostic to the target GNN architecture and training process. In other words, it only assumes black-box access to the target GNN model. 
    \item We develop custom optimizations to ensure the good performance of the underlying RL agent (Secs.~\ref{sec:challenge_and_solution_1} and~\ref{sec:challenge_and_solution_2}).
    \item Using a contextual Markov decision process formulation, we perform multi-task learning, enabling a single RL agent to successfully generate adversarial examples against GNNs used for four different classes of hardware security problems (Sec.~\ref{sec:challenge_and_solution_3}).
    \item Our results in Sec.~\ref{sec:results} demonstrate that \myname{} thwarts, i.e., successfully generates adversarial examples for all circuits against, 
    GNN-based techniques for four classes of hardware security problems: IP piracy detection, detecting/localizing HTs, reverse engineering circuits, and breaking hardware obfuscation techniques (see Table~\ref{tab:overview_table}).
    \item We demonstrate the ramifications of~\myname{}-generated adversarial examples through two practical, real-world attacks: fooling IP piracy detector on the \texttt{MIPS} and \texttt{IBEX} processors, and devising an HT that successfully leaks \texttt{AES} secret key by fooling an HT localization technique (Sec.~\ref{sec:case_studies}).
\end{itemize}

%% file: texfiles/background.tex
\section{Background}\label{sec:background}

\subsection{Graph Neural Networks}
Graph neural networks (GNNs) have emerged as a powerful framework for analyzing and modeling structured data represented by graphs. Typically, GNNs used in classification tasks learn representations of nodes in a graph by repeatedly aggregating and transforming the information (i.e., features) from their neighbor nodes. After a fixed number of aggregation iterations, the aggregated features are reduced by taking their sums, averages, or maximums. The reduced outputs are passed to a classifier (e.g., a two-layer fully-connected network) for final classification. GNNs have proven successful in protein folding~\cite{strokach2020fast}, social networks~\cite{fan2019graph}, and combinatorial optimization~\cite{cappart2023combinatorial}, among other fields.

\subsection{GNNs' Applications in Hardware Security}
Researchers have developed several GNN-based techniques for hardware security. Our selection of target GNN-based techniques ranges from the earliest techniques with high popularity, GNN4TJ~\cite{GNN4TJ} and GNN4IP~\cite{GNN4IP}, to the most recent one, TrojanSAINT~\cite{TrojanSAINT_ISCAS}. We also select other GNNs, OMLA~\cite{OMLA} and GNNRE~\cite{GNNRE}, based on their high success rates and variety of underlying GNN frameworks (see Table~\ref{tab:overview_table}). Overall, our selection represents a set of GNN-based techniques that use different frameworks and have demonstrated good performance for a variety of problems in hardware security.

\noindent\textbf{GNN for Intellectual Property (IP) Piracy Detection.} IP Piracy refers to the theft of the design IP by an adversary to develop competing devices without incurring research and development costs. GNN4IP is a GNN-based IP piracy detection technique that evaluates the similarity between two circuits~\cite{GNN4IP}. 
It converts the two circuits into a graph representation and uses GCNs to obtain graph embeddings, which are passed through a fully connected layer that outputs a similarity score. If the similarity between the original and test circuits is high, GNN4IP flags the test circuit as pirated.

\noindent\textbf{GNN for Hardware Trojan (HT) Detection.} HTs are malicious modifications an adversary makes to disrupt the original functionality. 
GNN4TJ is a GNN-based detection technique targeting HTs inserted in third-party IPs~\cite{GNN4TJ}. Similar to GNN4IP, GNN4TJ converts a given circuit into a graph representation, which is passed through a GCN, resulting in graph embedding. A fully-connected layer decides if the circuit has an HT or not using the embedding. 

\noindent\textbf{GNN for Localizing HTs.} TrojanSAINT is a GNN-based HT localization technique~\cite{TrojanSAINT_ISCAS}. Similar to other techniques, given a circuit, TrojanSAINT operates on its graph representation and classifies each node as HT-free or HT-infested.

\noindent\textbf{GNN for Reverse Engineering.} Reverse engineering refers to identifying different parts of a circuit with the intent of duplicating them. 
Similar to previous techniques, GNN-RE converts circuits into graphs and uses GCN layers followed by a fully connected network to classify gates into different modules/classes such as adders, multipliers, control logic, etc~\cite{GNNRE}. GNN-RE achieves an average accuracy of $98.82\%$ on benchmark circuits~\cite{GNNRE}.

\noindent\textbf{GNN for Hardware Obfuscation.} Hardware obfuscation is a design-for-trust scheme that promises protection throughout the IC supply chain by obfuscating certain circuit regions using key-controlled gates. OMLA is a GNN-based attack on hardware obfuscation that uses the structural information around the key-controlled gates to recover the correct key bits, thus breaking the security offered by the obfuscation~\cite{OMLA}.
OMLA achieves a high key-prediction accuracy (as high as >$90\%$), outperforming prior works on all benchmarks~\cite{OMLA}.

The aforementioned techniques report very high success rates and show great potential in addressing their respective hardware security problem. However, they all lack in terms of a crucial aspect: thorough evaluation of robustness to adversarial examples. Evaluation of these GNN-based techniques, and any machine learning technique in general, against adversarial examples is absolutely essential because adversarial examples can have drastic impacts. For instance,~\cite{eykholt2018robust} devised adversarial examples against image classification neural networks, resulting in misclassification of the ``STOP'' sign as a speed limit sign, which can cause a disaster in self-driving vehicles that use such image classifiers. The need for adversarial evaluation is especially pressing in these GNNs that target security applications. We develop~\myname{} as a framework to red-team these GNNs.

\subsection{Reinforcement Learning}
RL is a powerful framework in the field of artificial intelligence that enables an agent to learn and make sequential decisions in dynamic environments through interaction and feedback. Rooted in the concept of learning from rewards, RL employs an iterative process where an agent interacts with an environment, receives feedback in the form of rewards, and adjusts its behavior to maximize cumulative rewards over time.
By learning an optimal policy (a function that maps state-action pairs to probabilities of selecting a particular action in a given state), the RL agent aims to make informed decisions in different states to maximize its long-term rewards.
This learning paradigm is particularly well-suited for solving Markov decision processes (MDPs), which are mathematical models used to represent decision-making problems with sequential interactions.
RL has demonstrated remarkable success in various domains, including robotics, game playing, and resource management~\cite{lillicrap2015continuous,mnih2013playing,mao2016resource}.

%% file: texfiles/threat_model.tex
\section{Threat Model}\label{sec:threat_model}
We consider a standard and widely-used threat model of adversarial example generation~\cite{jin2021adversarial,wu2023kenku,li2023black,dai2018adversarial,mu2021hard,zheng2021black}. To that end, we make the following assumptions about the attacker.

\noindent\textbf{Attacker's Capacity.} The adversarial attack happens after the model has been trained. The model is fixed and the adversary cannot change the model parameters or structure. In particular, the attacker cannot poison the model and inject backdoors in the model. 

\noindent\textbf{Attacker's Abilities.} The attacker can introduce arbitrary perturbations, albeit those perturbations cannot change the functionality of the circuit, and he/she cannot violate circuit design rules. These perturbations include, but are not limited to, any combination of adding/deleting edges, injecting nodes, etc. as long as the final perturbed circuit maintains the original functionality and does not violate circuit design rules.

\noindent\textbf{Attacker's Knowledge.} Attacker's knowledge refers to the amount of information known to the attacker about the model he/she aims to attack. We assume a black-box setting. The attacker does not have access to the model's parameters or training labels. He/she can only perform black-box queries for output scores or labels.

\noindent\textbf{Attacker's Goal.} The attacker aims to generate input samples (i.e., circuits) that result in misclassification by the target GNN model. For instance, when the target model is GNN4IP~\cite{GNN4IP} (GNN-based technique for detecting IP piracy between two input circuits), given any original circuit, the attacker aims to create a pirated version of the circuit (by perturbing the original circuit) so that GNN4IP is fooled into classifying the perturbed circuit as ``not pirated''.

\noindent \textbf{Note.} The objective of this work is not to propose a new technique for inserting/detecting hardware Trojans (HTs), detecting/evading IP piracy, or reverse engineering. In other words,~\myname{} 
is not an attack or a defense on hardware security techniques. Rather, as mentioned in Table~\ref{tab:overview_table}, it is an attack on GNNs used in hardware security, be they used for performing malicious acts or for benevolent acts.

%% file: texfiles/methodology.tex
\section{Methodology}\label{sec:methodology}

We now demonstrate how prior works have generated adversarial examples against general GNNs using perturbations and how these techniques are not applicable in our case with circuits (Sec.~\ref{sec:challenges_for_adversarial_example_generation}). Then, using GNN4IP~\cite{GNN4IP} (a GNN-based technique for detecting IP piracy) as a representative example, we devise a novel approach to fool GNNs in hardware security
by formulating the problem of finding adversarial examples as an RL problem (Sec.~\ref{sec:preliminary_formulation}). However, this preliminary formulation suffers from generalization and performance issues which we overcome in Secs.~\ref{sec:challenge_and_solution_1} and ~\ref{sec:challenge_and_solution_2}. Additionally, for practicality and scalability, we devise a formulation that allows a single RL agent to generate successful adversarial examples against all GNNs (Sec.~\ref{sec:challenge_and_solution_3}). Ultimately, we tie everything together in Sec.~\ref{sec:final_formulation}.

\subsection{Limitations of Existing Adversarial Example Generation Techniques}\label{sec:challenges_for_adversarial_example_generation}
Adversarial examples in the context of GNNs refer to inputs that are purposefully crafted to deceive the GNN's predictions. Typically, adversarial examples in GNNs are generated by introducing perturbations to the input graph data. These perturbations are carefully designed to exploit vulnerabilities or limitations in the GNN model, causing it to make incorrect predictions. 
Researchers have developed a variety of perturbation techniques. These perturbation techniques use one or more of the following four approaches: adding edges, deleting edges, injecting nodes, and modifying features.
However, such perturbations cannot be used for hardware security problems that operate on Boolean circuits because (i) these perturbations affect the functionality of the circuit, (ii) they may also lead to violations of circuit design rules, and/or (iii) they violate our threat model.

\begin{figure}[t]
    \centering
    \includegraphics[width=0.5\textwidth,trim={1.3cm 0.8cm 1.3cm 0.9cm},clip]{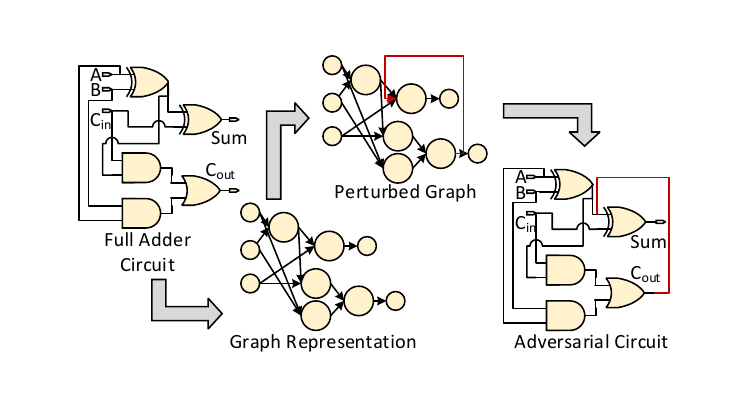}
    \smallerspacecaption
    \smallerspacecaption
    \smallerspacecaption
    \caption{Illustration of why existing adversarial example generation techniques are inappropriate for our case.
    }
    \label{fig:motivational_example}
    \smallerspacecaption
\end{figure}

\noindent\textbf{Example.} Figure~\ref{fig:motivational_example} uses a full adder circuit, its graph representation, an example perturbation, and the corresponding adversarial circuit to demonstrate why such perturbations cannot be used to generate adversarial examples for Boolean circuits. The perturbation adds an edge (shown in red) to the graph. However, doing so (i) changes the functionality of the circuit and (ii) has two drivers for a gate input, which violates circuit design rules. 
Similarly, the perturbation techniques that delete edges or inject nodes cannot be directly applied to GNNs used in hardware security. 
Moreover, modifying features is not applicable because although this perturbation technique would neither change the functionality of the circuit nor result in design rule violations, our threat model (Sec.~\ref{sec:threat_model}) prevents us from controlling the node features directly.

Since these widely used perturbation techniques against GNNs are not applicable to our case, we devise a new way to generate adversarial examples that satisfy the constraints of not altering the circuit's functionality, not violating design rule constraints, and not violating our threat model. Our perturbation involves a series of modifications to the gates and wires of the circuit, but for each modification, we ensure that the circuit's functionality remains unchanged and the design rule constraints are not violated. Hence, after applying this series of functionally equivalent modifications, we obtain the final perturbed circuit that is functionally equivalent to the original circuit but can still result in misclassification by the target GNN.

\subsection{Preliminary Formulation}\label{sec:preliminary_formulation}
As explained above, we must devise a series of perturbations that do not change the circuit's functionality but still result in misclassification. To achieve these functionality-preserving perturbations, we rely on primitive circuit transformations applied on circuits during the synthesis process, i.e., when a Verilog (or even a circuit) circuit description is translated (or re-translated) into a circuit.
Here, since we use the open-source synthesis tool ABC~\cite{ABC},\footnote{ABC is a widely-used open-source synthesis tool developed at UC Berkeley.} we use its primitive circuit transformations, such as \texttt{refactor}, \texttt{refactor -z}, \texttt{resub}, \texttt{balance}, etc., to cause functionality-preserving perturbations.
When applied to a circuit, these transformations change the circuit's structure but not the function. However, the order of application of these transformations affects the perturbed circuit and, hence, the success of the perturbation. Moreover, different circuits have different structures and thus require different transformations in different orders. So, to find the optimal sequence of transformations for a given circuit that results in misclassification by GNN4IP, we design an RL agent that finds the optimal policy for (i.e., solves) the MDP, denoted as a tuple $(\mathcal{S}, \mathcal{A}, P(s_{t+1}|a_t,s_t), R(s_t,a_t), \gamma)$, defined as follows:

\begin{itemize}[leftmargin=*]

\item \textbf{States $\mathcal{S}$} is the set of all possible values of the state vector. The state vector, $s_t$, that characterizes the circuit at time $t$, is defined as a list of pre-determined features in that circuit:
$s_t = [$\# inputs, \# outputs, \# gates, \# wires, \# AND gates, \# OR gates, \# NAND gates, \# NOR gates, \# INV gates, \# BUF gates, \# XOR gates, \# XNOR gates, \# other gates$]$, where ``other gates'' refers to all gate types not explicitly listed in the array, e.g., flip-flops.
We select these features since the resulting state vector captures information about the structure of the circuit, and it is closely related to the node features used in GNN4IP~\cite{GNN4IP}.

\item \textbf{Actions $\mathcal{A}$} is the set of the following functionality-preserving transformations in ABC: \{\texttt{refactor}, \texttt{rewrite}, \texttt{resub}, \texttt{balance}, \texttt{refactor -z}, \texttt{rewrite -z}, \texttt{resub -z}\}. We choose these transformations since they can produce different resynthesized circuits while preserving functionality~\cite{ABC}.
We also add the \texttt{no-op} (short for ``no operation'') action to this set to denote no action. Doing so allows the agent to not perturb more if the current perturbed circuit is sufficient to evade GNN4IP.
An individual action, $a_t$, is the transformation chosen by the agent at time $t$.

\item \textbf{State transition $P(s_{t+1}|a_t,s_t)$} is the probability that action $a_t$ in state $s_t$ leads to the state $s_{t+1}$. 
In our case, the chosen transformation (i.e., the action $a_t$) is provided to ABC, which applies the transformation to the current circuit (represented by state $s_t$), and results in the resynthesized circuit (represented by state $s_{t+1}$). Note that since this transformation is deterministic, the state transition is also deterministic:
    \[
    P(s_{t+1}|a_t,s_t) = 
    \begin{dcases}
    1, & \text{if } \text{ ABC}(s_t,a_t) = s_{t+1}\\
    0, & \text{otherwise}
    \end{dcases}
    \]
    
\item \textbf{Reward function $R(s_t,a_t) = r_t$} is equal to $\alpha$ ($>0$) if the next state is misclassified by GNN4IP as not pirated; it is $0$ otherwise.
    \[
    R(s_{t},a_t)= r_t =
    \begin{dcases}\label{eq:reward_eq_GNN4IP}
        \alpha, & \text{if } \text{ GNN4IP}(s_1,s_{t+1}) = \text{ not pirated}\\ 
        0,              & \text{otherwise} \tag{1}
    \end{dcases}
    \]
Here, $s_1$ is the initial state, i.e., the original circuit we wish to pirate, and $\text{GNN4IP}(N_A,N_B)$ is the trained GNN-based IP-piracy detector function that takes as input two circuits, $N_A$ and $N_B$, and returns ``pirated'' if it determines that circuit $N_A$ is pirated from $N_B$ or vice versa (since the function is symmetric) and returns ``not pirated'' otherwise. The reward is designed so that the agent tries to successfully evade detection by GNN4IP with the smallest number of perturbations.
    
\item \textbf{Discount factor $\gamma$} $(0 \leq \gamma \leq 1)$ indicates the importance of future rewards relative to the current reward.
\end{itemize}

The initial state $s_1$ is a randomly picked (from the set of all circuits GNN4IP is trained with) original circuit 
that we wish to pirate and fool GNN4IP with. 
At each step, $t$, the agent in state $s_t$ chooses an action $a_t$, arrives in the next state $s_{t+1}$ according to the state transition rules, and receives a reward $r_{t}$. 
This cycle of state, action, reward, and next state is repeated $T$ (a pre-determined finite number) times, constituting one \textit{episode}.
At the end of each episode, the agent's state reflects the final perturbed circuit.
We train our agent using the Proximal Policy Optimization
algorithm with default parameters unless specified otherwise~\cite{schulman2017proximal}.

Our experiments indicate that this preliminary agent performs well for some circuits but not for most of the circuits on which GNN4IP is trained. We analyzed the agent in greater detail and discovered some challenges faced by this preliminary formulation, which are explained and addressed next.

\begin{table}[t]
\caption{New actions based on allowed/prohibited standard cells. AND$x$ indicates an $x$-input AND gate. \CIRCLE~indicates allowed standard cells, and \Circle~indicates standard cells that are prohibited for that action.}
\label{tab:limitation_solution_1_actions}
\smallerspacecaption
\resizebox{0.48\textwidth}{!}{%
\begin{tabular}{ccccccc}
\toprule
Action & \begin{tabular}[c]{@{}c@{}}AND2,\\ OR2\end{tabular} & \begin{tabular}[c]{@{}c@{}}NAND2,\\ NOR2\end{tabular}& \begin{tabular}[c]{@{}c@{}}AND$x$, OR$x$\\ ($x\geq$3)\end{tabular} & \begin{tabular}[c]{@{}c@{}}NAND$x$, NOR$x$\\ ($x\geq$3)\end{tabular} & \begin{tabular}[c]{@{}c@{}}XOR,\\ XNOR\end{tabular} & \begin{tabular}[c]{@{}c@{}}INV,\\ BUF\end{tabular}\\
 \midrule
$a^1$ & \CIRCLE & \CIRCLE & \Circle & \Circle & \CIRCLE & \CIRCLE \\ \midrule
$a^2$ & \CIRCLE & \CIRCLE & \CIRCLE & \CIRCLE & \CIRCLE & \CIRCLE \\ \midrule
$a^3$ & \Circle & \CIRCLE & \Circle & \Circle & \Circle & \CIRCLE \\ \midrule
$a^4$ & \Circle & \CIRCLE & \Circle & \CIRCLE & \Circle & \CIRCLE \\ \midrule
$a^5$ & \Circle & \CIRCLE & \Circle & \Circle & \CIRCLE & \CIRCLE \\ \midrule
$a^6$ & \Circle & \CIRCLE & \Circle & \CIRCLE & \CIRCLE & \CIRCLE \\ \midrule
$a^7$ & \CIRCLE & \Circle & \Circle & \Circle & \CIRCLE & \CIRCLE \\ \midrule
$a^8$ & \CIRCLE & \Circle & \CIRCLE & \Circle & \CIRCLE & \CIRCLE \\ \midrule
$a^9$ & \CIRCLE & \Circle & \Circle & \Circle & \Circle & \CIRCLE \\ \midrule
$a^{10}$ & \CIRCLE & \Circle & \CIRCLE & \Circle & \Circle & \CIRCLE \\ \bottomrule
\end{tabular}
}
\smallerspacecaption
\smallerspacecaption
\end{table}

\begin{figure}[t]
    \centering
    \includegraphics[width=0.48\textwidth,trim={0 0 0 0}, clip]{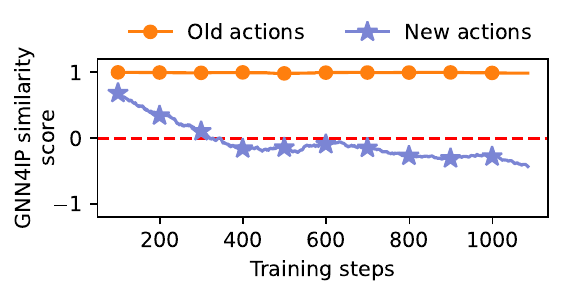}
    \smallerspacecaption
    \smallerspacecaption
    \smallerspacecaption
    \smallerspacecaption
    \caption{Comparison of the evolution of GNN4IP similarity scores with old and new actions as training progresses.}
    \label{fig:limitation_solution_1}
    \smallerspacecaption
    \smallerspacecaption
\end{figure}

\subsection{Effective and Generalizable Actions}\label{sec:challenge_and_solution_1}
\textbf{Challenge: Ineffective and Specific Actions.} The preliminary formulation relies only on the transformations available in the ABC synthesis tool to perturb the circuit. Those transformations (i.e., the actions defined above) have two issues. (i) They do not change the state significantly for several of the circuits.
Thus, GNN4IP easily detects the structural similarity between the original and the pirated circuits. 
(ii) They are specific to the ABC synthesis tool, resulting in virtually zero compatibility with other open-source as well as industrial standard commercial synthesis tools such as Synopsys Design Compiler~\cite{Design_Compiler} and Cadence Genus~\cite{Genus}.

\noindent\textbf{Solution \cc{1}:} 
To address this challenge, we devise novel, more effective (albeit still functionality-preserving) actions for the agent that are extremely likely to change the state, i.e., the node features of the pirated circuit. These novel actions are the 10 different gate type (also called ``standard cell'') selection strategies shown in Table~\ref{tab:limitation_solution_1_actions}. For instance, if action $a^1$ is chosen, the synthesis tool (i.e., ABC, Synopsys Design Compiler, or Cadence Genus) is allowed to use the 2-input AND (AND2 in the table), OR, NAND, and NOR standard cells (i.e., gates), and XOR, XNOR, INV, and BUF standard cells, but not allowed to use $3$ or higher input AND, OR, NAND, and NOR standard cells. So, if, in state $s_t$, action $a^1$ is chosen, $s_{t+1}$ will not contain 3-input AND, OR, NAND, and NOR standard cells.
Another advantage of these new actions based on standard cell selection strategies is that they are agnostic to the synthesis tool. All widely used synthesis tools, both open-source and commercial, are compatible with these new actions, unlike the actions defined in the preliminary formulation. This synthesis-tool-agnostic nature of the actions is essential for generating adversarial examples against some of the GNNs in hardware security, as explained later in Sec.~\ref{sec:results}.

Simply adding these $10$ new actions to the previously mentioned $8$ actions (\texttt{refactor}, \texttt{rewrite}, ..., etc.) would result in an extremely large action space ($(10\times8)^{T}$, where $T$ is the number of steps in an episode) for the agent. Hence, to reduce the action space, for each of the $T$ steps in an episode, we only let the agent choose one of the $10$ standard cell strategies and apply three fixed transformations (if using ABC): \texttt{rewrite}, \texttt{balance}, \texttt{refactor} (in that order). In other words, at each step, the action space of the agent is the set of $10$ functionality-preserving transformations shown in Table~\ref{tab:limitation_solution_1_actions} (and the \texttt{no-op} action for no change to the state; this allows the agent not to perturb more when the current perturbed circuit is sufficient to evade GNN4IP), and for each individual action from those, we apply the fixed transformations (if using ABC) \texttt{rewrite}, \texttt{balance}, and \texttt{refactor} in that order. 

Figure~\ref{fig:limitation_solution_1} compares the GNN4IP similarity scores as a function of first $\approx 1K$ training steps for $10$ \texttt{c432} benchmarks from the GNN4IP repository~\cite{HW2VEC_github}. Lower similarity score indicates successful attack. The dashed red line is GNN4IP's threshold for classifying a circuit as pirated.
The superiority of the agent's learning and performance is clearly visible: with the new actions, the agent quickly learns to generate adversarial circuits that fool GNN4IP (leading it to classify pirated circuits as not pirated), whereas with the old actions, the agent is unable to generate successful adversarial examples.

\subsection{Sparse Rewards for Faster Training}\label{sec:challenge_and_solution_2}
\textbf{Challenge: Unnecessary Reward Computations} Another challenge faced by the preliminary formulation is that it involves reward computation at each step. Reward computation requires querying the trained GNN4IP model with the updated state, $s_{t+1}$, to determine if it is classified as pirated or not. Since this involves loading the trained model, parsing the original and current circuits, and performing a forward pass of the GNN, it consumes at least a few seconds. Since RL agents typically need several thousands, if not tens of thousands of steps, to learn, this time-intensive reward computation slows the RL training process dramatically.

\noindent\textbf{Solution \cc{2}:} To reduce training time,
we employ the strategy of computing rewards only at the end of the episode instead of at each step of the episode.
Doing so reduces the frequency of reward computation, leading to less time per episode during training. Note that computing rewards at the end of the episode instead of at each step can affect the performance of the agent, i.e., it may lead to sub-optimal convergence. However, our results show that our agent still converges to an effective policy, i.e., it learns to generate successful adversarial examples. Table~\ref{tab:limitation_solution_2} shows 
the two ways to provide rewards (at each step and at the end of each episode) and their impacts on the training rates and the percentage of successful episodes 
against GNN4IP. It is evident that the sparse reward computation increases the rate dramatically while actually improving the percentage of successful episodes.
Hence, we use sparse rewards (i.e., at the end of each episode) to train our RL agent.

\begin{table}[tb]
\caption{Comparison of training rates for the reward methods: at each step vs. at end of episode.}
\label{tab:limitation_solution_2}
\smallerspacecaption
\centering
\resizebox{0.48\textwidth}{!}{%
\begin{tabular}{cccc} \toprule
\multirow{2}{*}{Method} & \multirow{2}{*}{\begin{tabular}[c]{@{}c@{}}\% of successful \\ episodes\end{tabular}} & \multicolumn{2}{c}{Rate}  \\ \cmidrule(lr){3-4}
& & (steps/min) & (eps./min) \\ \midrule
Reward at each step & $77$ & $18.13$ & $3.46$\\ 
Reward at end of episode & $89$ & $67.73$ & $13.33$\\ 
Improvement & $1.15\times$ & $3.73\times$ & $3.85\times$\\ \bottomrule
\end{tabular}%
}
\smallerspacecaption
\smallerspacecaption
\end{table}

\subsection{Multi-Task Learning}\label{sec:challenge_and_solution_3}
\noindent \textbf{Challenge: MDP Specific to one GNN.} So far, we formulated an MDP, that when solved by an RL agent, yields adversarial examples against GNN4IP. However, this MDP is specific to GNN4IP. If we wish to target other GNN techniques, we would need to devise separate MDPs, each with their separate RL agents. In other words, we would have different RL agents to learn different tasks, i.e, generate adversarial examples against different GNNs. However, training separate RL agents for different tasks is not ideal because each RL agent would be independent and would require training from scratch instead of learning knowledge common among different tasks. This would result in a large runtime to generate adversarial examples against all the GNNs, limiting the scalability of our technique. To overcome this challenge, we need to devise a single RL agent that learns different tasks, i.e., generates successful adversarial examples against all GNNs.

\noindent \textbf{Solution \cc{3}:} We devise a contextual Markov decision process (CMDP) formulation that can enable multi-task learning by a single RL agent. A CMDP is denoted as a tuple $(\mathcal{C}, \mathcal{S}, \mathcal{A}, \mathcal{M}(c))$, where $\mathcal{C}$ is called the context space, $\mathcal{S}$ is the state space, $\mathcal{A}$ is the action space, and $\mathcal{M}$ is a function mapping any context $c \in \mathcal{C}$ to an MDP $\mathcal{M}(c) = (\mathcal{S}, \mathcal{A}, P^c(s_{t+1}|a_t,s_t), R^c(s_t,a_t), \gamma^c )$. In other words, given a context $c\in \mathcal{C}$, the CMDP reduces to a regular MDP specific to that context. A key feature required to formulate a CMDP is that the state and action spaces of all the constituent MDPs need to be the same. Since our state and action formulations from Sec.~\ref{sec:challenge_and_solution_1} are agnostic to the underlying GNN, we can formulate the CMDP against all GNNs by designing different appropriate reward functions for the different GNNs. In other words, we can construct a CMDP that encompasses MDPs against all GNNs and then train a single RL agent that finds the optimal policy for the CMDP, and hence for all its constituent MDPs. Next, we formulate this CMDP.

\begin{figure*}[t]
    \centering
    \includegraphics[width=\textwidth,trim={0 0 0 0}, clip]{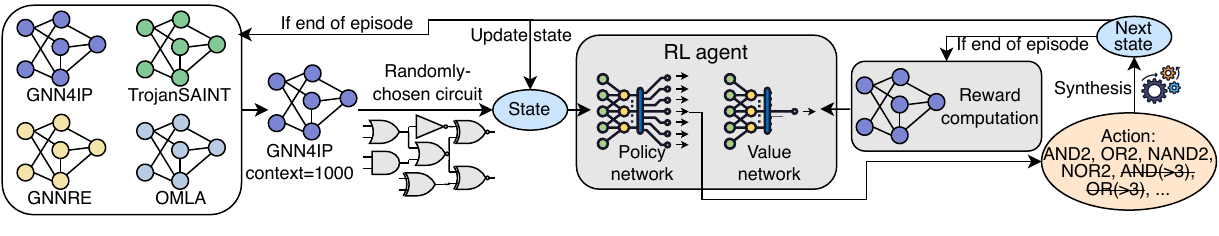}
    \smallerspacecaption
    \smallerspacecaption
    \smallerspacecaption
    \smallerspacecaption
    \caption{Final~\myname~architecture.}
    \label{fig:RL_architecture}
    \smallerspacecaption
    \smallerspacecaption
\end{figure*}

\begin{itemize}[leftmargin=*]
\item \textbf{The context space $\mathcal{C}$} is the set of one-hot encoded binary strings, one for each GNN we target. Since we target four GNNs (GNN4IP, TrojanSAINT, GNN-RE, and OMLA), $\mathcal{C} = \{1000, 0100, 0010, 0001\}$, with $1000$ reducing the CMDP to the MDP for GNN4IP, and so on.
\item \textbf{States $\mathcal{S}$} of the CMDP are as defined in Sec.~\ref{sec:preliminary_formulation}. 
\item \textbf{Actions $\mathcal{A}$} of the CMDP are as defined in Table~\ref{tab:limitation_solution_1_actions}.
\item \textbf{State transitions $P^c(s_{t+1}|a_t,s_t)$} of the CMDP are the as defined in Sec.~\ref{sec:preliminary_formulation} for all $c \in \mathcal{C}$. 
\item \textbf{Reward functions $R^c(s_t,a_t)$} for the constituent MDPs of the CMDP are defined separately as follows:

\textbf{TrojanSAINT} is a GNN-based hardware Trojan (HT) localization technique. Given a set of nodes (i.e., gates in a circuit), it classifies each node as HT-free or HT-infested, which helps determine the location of an HT in the circuit. To generate adversarial examples against TrojanSAINT, we 
design the 
end-of-episode reward function (hence subscript $T$) as:
\begin{equation}\label{eq:reward_eq_TrojanSAINT}
    R(s_{T},a_T)= r_T =  1 - \alpha_{TS}(s_{T+1}) \tag{2}
\end{equation}
Here, $\alpha_{TS}(N)$ is the performance of TrojanSAINT on the input HT-infested circuit, $N$, measured according to~\cite{TrojanSAINT_ISCAS} as the average of true positive and true negative rates.

\textbf{GNN-RE} classifies gates in a circuit into different modules (adders, subtractors, comparators, multipliers, and control logic). To generate adversarial examples against GNN-RE, we design the reward function as:
\begin{equation}\label{eq:reward_eq_GNNRE}
    R(s_{T},a_T)= r_T =  1 - \alpha_{RE}(s_{T+1}) \tag{3}
\end{equation}
Here, $\alpha_{RE}(N)$ is the accuracy of the trained GNN-RE classifier that takes as input a circuit, $N$, and returns the labels (``adder'', ``subtractor'', ``comparator'', ``multiplier'', or ``control logic'') for the nodes in $N$.

Since \textbf{OMLA} uses GNNs to predict the key bits used to obfuscate the circuit, the adversarial examples are designed to result in poor classification accuracy. To this end, we design the reward function as:
\begin{equation}\label{eq:reward_eq_OMLA}
    R(s_{T},a_T)= r_T =  e^{-5|0.5-\alpha_{OMLA}(s_{T+1})|} \tag{4}
\end{equation}
Here, $\alpha_{OMLA}(N)$ is the key prediction accuracy of the trained OMLA GNN that takes as input an obfuscated circuit, $N$, and returns the predicted key bits ($0$ or $1$ for the key gates (i.e., the obfuscation gates that take key bits as inputs) in $N$. The reward is designed to provide marginally increasing returns as OMLA's accuracy drops closer to 0.5, i.e., it performs no better than a random guess.

In summary, the reward functions for MDPs are designed to generate adversarial examples 
so that the corresponding GNN yields low accuracy or a high misclassification rate. Finally, the reward function for the CMDP is a congregation of the individual MDPs' rewards:
\[
    R^c(s_{t},a_t) =
    \begin{dcases}
        Eq. (\ref{eq:reward_eq_GNN4IP}), & \text{if } $c=1000$\\
        Eq. (\ref{eq:reward_eq_TrojanSAINT}), & \text{if } $c=0100$\\
        Eq. (\ref{eq:reward_eq_GNNRE}), & \text{if } $c=0010$\\
        Eq. (\ref{eq:reward_eq_OMLA}), & \text{if } $c=0001$
    \end{dcases}
    \]
\item \textbf{Discount factors $\gamma^c$} for the CMDP are as defined in Sec.~\ref{sec:preliminary_formulation} for all $c \in \mathcal{C}$.
\end{itemize}
We solve this CMDP using a single RL agent, thus performing multi-task learning. Next, we unify all three solutions we developed and describe the final RL formulation that generates adversarial examples using \cc{1} effective and generalizable actions, \cc{2} sparse rewards for achieving faster training, and \cc{3} CMDP for multi-task learning.

\begin{figure*}[t]
    \centering
    \includegraphics[width=\textwidth,trim={0 0.2cm 0 0.2cm},clip]{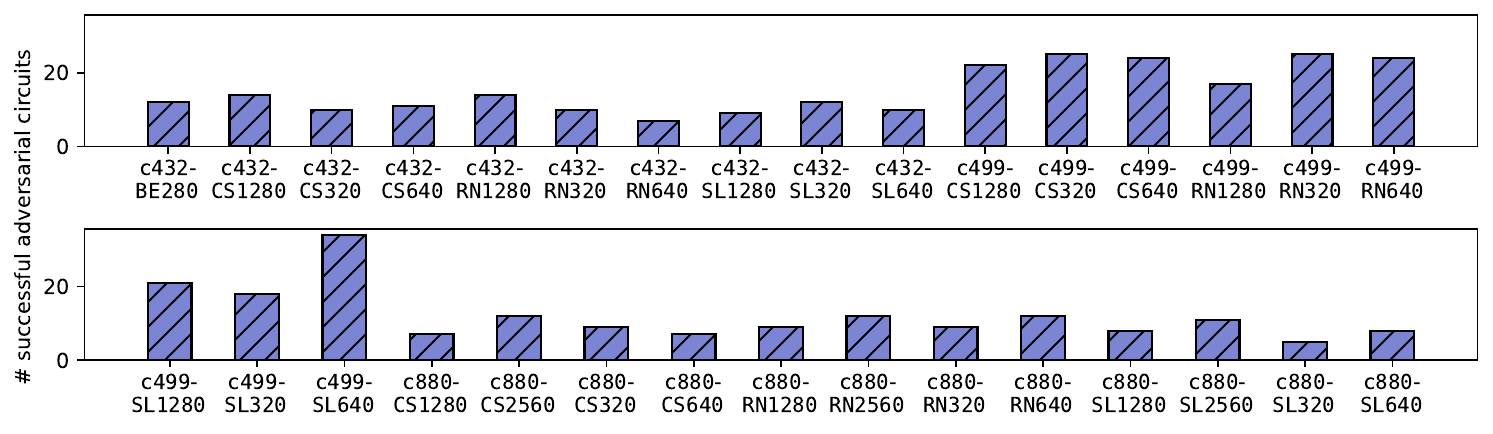}
    \smallerspacecaption
    \smallerspacecaption
    \smallerspacecaption
    \caption{Number of successful \myname{}-generated adversarial circuits against GNN4IP (higher values: better attack).
    }
    \label{fig:GNN4IP_adv_cnt}
\end{figure*}

\begin{figure*}[t]
    \centering
    \includegraphics[width=\textwidth,trim={0 0.2cm 0 0.2cm},clip]{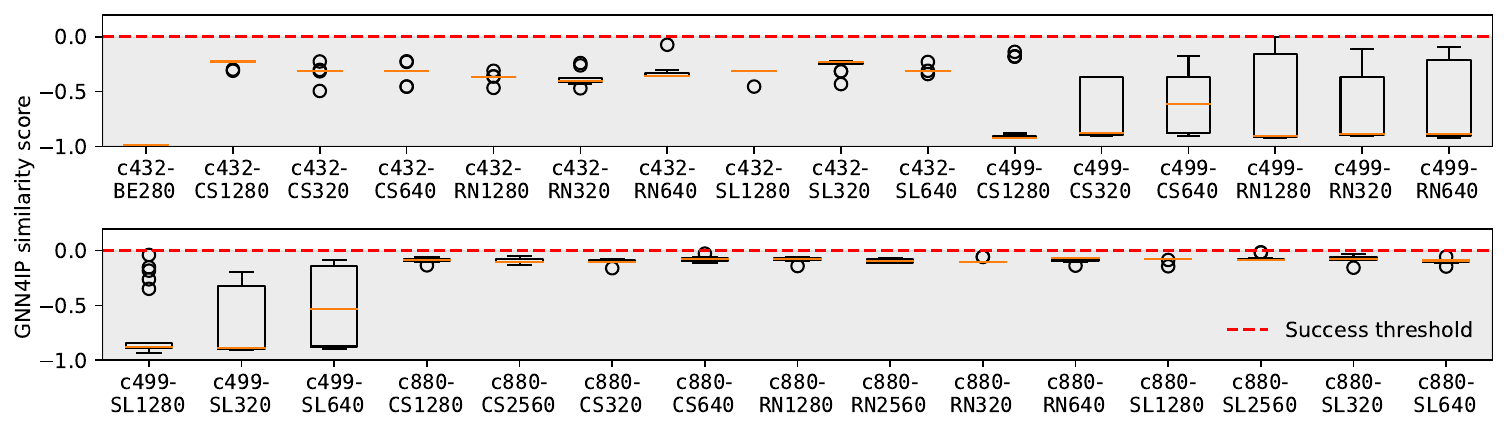}
    \smallerspacecaption
    \smallerspacecaption
    \smallerspacecaption
    \caption{Distribution of GNN4IP similarity scores for~\myname{}-generated adversarial circuits (lower values: better attack).
    }
    \label{fig:GNN4IP_sim_scores}
    \smallerspacecaption
\end{figure*}
\subsection{Final Formulation}\label{sec:final_formulation}
Figure~\ref{fig:RL_architecture} illustrates the final architecture of our RL agent against GNNs in hardware security. For each episode, the agent starts with a randomly picked circuit for a randomly picked target GNN (GNN4IP in the figure) and takes an action according to the policy parameterized by a neural network.\footnote{In addition to the policy neural network, the agent also contains a value neural network that predicts the expected reward for a given state-action pair, which is used to train the RL agent.}
Based on the action, a synthesis recipe is created, which is compiled using the appropriate synthesis tool (ABC/Synopsys Design Compiler/Cadence Genus) to generate the next state of the agent. More specifically, we use the open-source ABC tool~\cite{ABC} when working with GNN4IP and OMLA and the industry-standard Synopsys Design Compiler~\cite{Design_Compiler} when working with TrojanSAINT and GNN-RE since the latter GNNs require gate names to be preserved after synthesis for correct labeling, which is not supported in ABC.\footnote{Note that \myname{} is compatible with all synthesis tools.} Then, the agent chooses another action, and so on. This cycle is repeated $T$ times, which constitutes an episode. At the end of the episode, the final state is evaluated using the chosen GNN for that episode to produce a reward for the agent. After a fixed-size batch of episodes, the Proximal Policy Optimization (PPO) algorithm translates the rewards into losses, which are used by the Adam optimizer~\cite{kingma2014adam} to update the parameters of the neural networks that make up the agent. After several such batches of updates to the parameters, the rewards saturate, and the neural networks converge, resulting in the agent learning an optimal or a near-optimal policy to generate successful adversarial examples against all target GNNs. 
Next, we demonstrate the efficacy of this multi-task RL agent in generating adversarial examples.

%% file: texfiles/results.tex
\section{Results}\label{sec:results}

\subsection{Experimental Setup}
\label{sec:setup}
We implemented~\myname~using \textit{PyTorch 1.12} and \textit{stable-baselines3} and trained it using $16$ cores in a Linux machine with
a \textit{Dual AMD EPYC 7443} processor with a 256GB RAM.
We implemented custom parsers and glue scripts in \textit{Python3}. 
We used the Proximal Policy Optimization algorithm~\cite{schulman2017proximal} for training the RL agent.
We used a two-layered, $64\times64$ fully-connected neural network with Tanh activation function for our policy and value networks.
We selected the reward parameter, $\alpha$ for GNN4IP in Eq.~(\ref{eq:reward_eq_GNN4IP}), as $1$ to have it on the same scale as the other rewards (Eqs.~(\ref{eq:reward_eq_TrojanSAINT}), (\ref{eq:reward_eq_GNNRE}), and~(\ref{eq:reward_eq_OMLA})).
Our MDP formulation for identifying perturbations follows the state ($s$) $\rightarrow$ action ($a$) $\rightarrow$ next state $\rightarrow$ reward ($r$) flow. For GNN4IP and OMLA, we set $T$, the episode length, as $5$, so there is a five-step evolution of state: $s_0 \rightarrow a_0 \rightarrow s_1 \ldots \rightarrow s_5 \rightarrow r$. Whereas for TrojanSAINT and GNN-RE, we set $T$ as $1$, so there is a one-step evolution of state: $s_0 \rightarrow a_0 \rightarrow s_1 \rightarrow r$, meaning that there are two states in the flow. This is not supervised learning because $T$ must be $0$ for supervised learning, i.e., no state evolution. Moreover, we observed that setting $T$ as $1$ for TrojanSAINT and GNN-RE reduced the runtime and was still sufficient to generate successful adversarial examples using Synopsys Design Compiler.

\begin{figure*}[t]
    \centering
    \includegraphics[width=\textwidth,trim={0 0.2cm 0 0.2cm},clip]{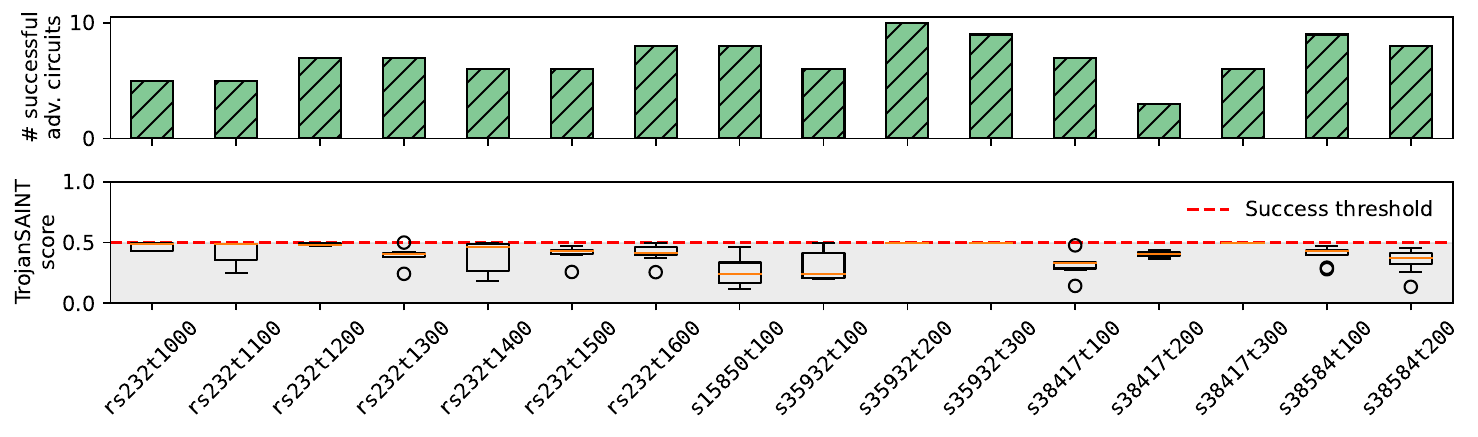}
    \smallerspacecaption
    \smallerspacecaption
    \smallerspacecaption
    \caption{Top: Number of successful \myname{}-generated adversarial circuits against TrojanSAINT (higher values: better attack). Bottom: Distribution of TrojanSAINT's scores for those adversarial circuits (lower values: better attack).
    }
    \label{fig:TrojanSAINT_adv_cnt_and_score}
    \smallerspacecaption
    \smallerspacecaption
\end{figure*}

\subsection{Success Against GNN4IP~\cite{GNN4IP}}\label{sec:results_GNN4IP}
We obtained the GNN4IP code and a dataset of $31$ different circuits from the GNN4IP repository~\cite{HW2VEC_github}. Using our RL agent, we generated adversarial examples for each of these $31$ circuits. Since GNN4IP detects IP piracy between two circuits, a successful adversarial example should fool GNN4IP into classifying a pirated circuit as not pirated, i.e., assign it a similarity score $<0$.  
Another important thing to note is that we generate adversarial examples by perturbing circuits from the training set itself, i.e., circuits that GNN4IP has seen during training. This is a more difficult setting for our attack than the typical setting where one perturbs a circuit from the testing set, i.e., circuits that GNN4IP has not seen before. We follow this difficult setting to showcase the exceptional capability of \myname{} in generating successful adversarial examples.

Figure~\ref{fig:GNN4IP_adv_cnt} shows the number of successful adversarial circuits found by our RL agent for each of the $31$ circuits from GNN4IP's training set. As the results show,\textbf{~\myname{} easily generates many successful adversarial circuits against GNN4IP}.
Figure~\ref{fig:GNN4IP_sim_scores} presents further analysis of our adversarial circuits vis-a-vis the distribution of the GNN4IP similarity scores for those circuits. The figure demonstrates that for most successful adversarial circuits, GNN4IP's similarity score is significantly less than $0$ even though all those adversarial circuits are actually pirated from the original circuits.

\subsection{Success Against TrojanSAINT~\cite{TrojanSAINT_ISCAS}}\label{sec:results_against_trojansaint}
We obtained the TrojanSAINT code from the TrojanSAINT repository~\cite{TrojanSAINT_github}. Moreover, following TrojanSAINT~\cite{TrojanSAINT_ISCAS}, we used $16$ HT-infested circuits 
from the TrustHub suite~\cite{TrustHub}. The TrustHub suite is a repository of many real-world circuits with various HTs that cause denial-of-service, degradation in performance, leak secret keys, etc.
For each of the $16$ circuits, we train a separate GNN as is done in TrojanSAINT~\cite{TrojanSAINT_ISCAS}. 
Then, we use the single \myname{} agent to generate adversarial circuits against all $17$ trained GNNs.
Since TrojanSAINT performs binary classification, we define an adversarial circuit, i.e., a perturbed HT-infested circuit, as successful if TrojanSAINT's score (average of true positive rate and true negative rate) is below $0.5$, i.e., $50\%$. 

Figure~\ref{fig:TrojanSAINT_adv_cnt_and_score} shows the number of successful \myname{}-generated adversarial circuits (top) and the TrojanSAINT score distribution (bottom) for them. We observe that even though \myname{} is not trained separately for each of the $17$ different TrojanSAINT GNNs, it easily generates plenty of successful adversarial circuits against TrojanSAINT. Moreover, like GNN4IP, \textbf{all successful adversarial circuits result in poor TrojanSAINT performance}.

\begin{figure*}[t]
    \centering
    \includegraphics[width=\textwidth, trim={0 0.2cm 0 0.2cm},clip]{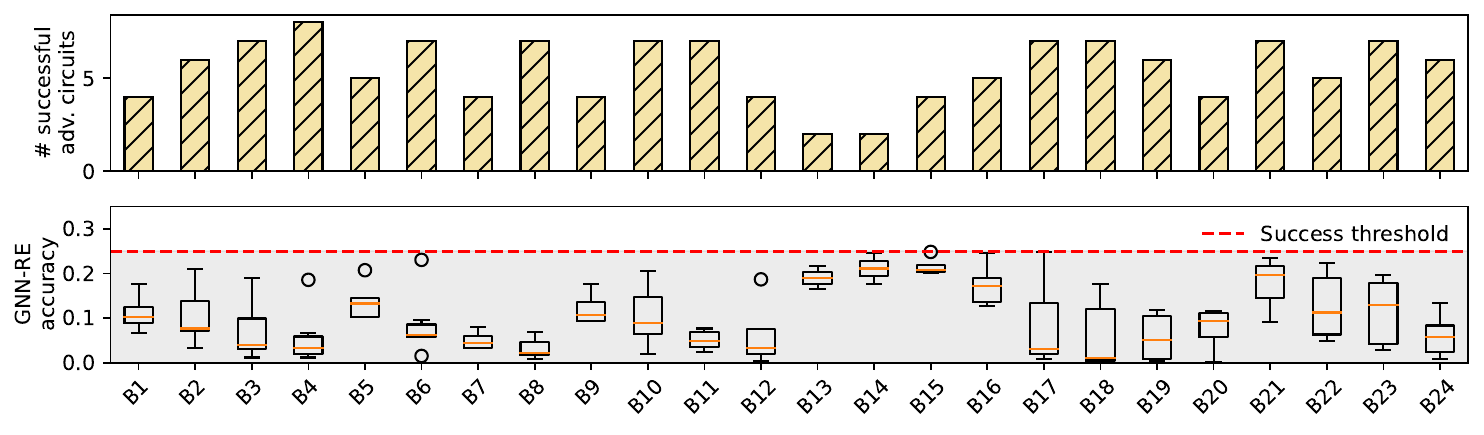}
    \smallerspacecaption
    \smallerspacecaption
    \smallerspacecaption
    \caption{Top: Number of \myname{}-generated successful adversarial circuits against GNN-RE (higher values: better attack) Bottom: Distribution of GNN-RE's accuracy for those adversarial circuits (lower values: better attack).
    }
    \label{fig:GNNRE_adv_cnt_and_accuracy}
    \smallerspacecaption
\end{figure*}

\begin{table*}[t]
\centering
\caption{Name encoding for the circuits in GNN-RE repository~\cite{GNNRE_github}. \CIRCLE{} and \Circle{} indicate the presence and absence of that substring in the name of the circuit benchmark, respectively. For instance, \texttt{B7} corresponds to the ``add\_mul\_combine\_16\_bit.v'' circuit from the GNN-RE repository~\cite{GNNRE_github}.}
\smallerspacecaption
\label{tab:GNNRE_benchmarks}
\resizebox{\textwidth}{!}{
\begin{tabular}{ccccccccccccccccccccccccc}
\toprule
 & \texttt{B1} & \texttt{B2} & \texttt{B3} & \texttt{B4} & \texttt{B5} & \texttt{B6} & \texttt{B7} & \texttt{B8} & \texttt{B9} & \texttt{B10} & \texttt{B11} & \texttt{B12} & \texttt{B13} & \texttt{B14} & \texttt{B15} & \texttt{B16} & \texttt{B17} & \texttt{B18} & \texttt{B19} & \texttt{B20} & \texttt{B21} & \texttt{B22} & \texttt{B23} & \texttt{B24} \\
 \midrule
Add & \CIRCLE & \CIRCLE & \CIRCLE & \CIRCLE & \CIRCLE & \CIRCLE & \CIRCLE & \CIRCLE & \CIRCLE & \CIRCLE & \CIRCLE & \CIRCLE & \CIRCLE & \CIRCLE & \CIRCLE & \CIRCLE & \CIRCLE & \CIRCLE & \CIRCLE & \CIRCLE & \CIRCLE & \CIRCLE & \CIRCLE & \CIRCLE \\
\midrule
Mul & \CIRCLE & \CIRCLE & \CIRCLE & \CIRCLE & \CIRCLE & \CIRCLE & \CIRCLE & \CIRCLE & \CIRCLE & \CIRCLE & \CIRCLE & \CIRCLE & \CIRCLE & \CIRCLE & \CIRCLE & \CIRCLE & \CIRCLE & \CIRCLE & \CIRCLE & \CIRCLE & \CIRCLE & \CIRCLE & \CIRCLE & \CIRCLE \\
\midrule
Sub & \Circle & \Circle & \Circle & \Circle & \Circle & \Circle & \Circle & \Circle & \Circle & \Circle & \Circle & \Circle & \CIRCLE & \CIRCLE & \CIRCLE & \CIRCLE & \Circle & \Circle & \Circle & \Circle & \CIRCLE & \CIRCLE & \CIRCLE & \CIRCLE \\
\midrule
Cmb & \Circle & \Circle & \Circle & \Circle & \CIRCLE & \CIRCLE & \CIRCLE & \CIRCLE & \Circle & \Circle & \Circle & \Circle & \Circle & \Circle & \Circle & \Circle & \Circle & \Circle & \Circle & \Circle & \Circle & \Circle & \Circle & \\
\midrule
Cmp & \Circle & \Circle & \Circle & \Circle & \Circle & \Circle & \Circle & \Circle & \CIRCLE & \CIRCLE & \CIRCLE & \CIRCLE & \CIRCLE & \CIRCLE & \CIRCLE & \CIRCLE & \Circle & \Circle & \Circle & \Circle & \Circle & \Circle & \Circle & \Circle \\
\midrule
Mix & \Circle & \Circle & \Circle & \Circle & \Circle & \Circle & \Circle & \Circle & \Circle & \Circle & \Circle & \Circle & \Circle & \Circle & \Circle & \Circle & \CIRCLE & \CIRCLE & \CIRCLE & \CIRCLE & \Circle & \Circle & \Circle & \Circle\\
\midrule
\#Bits & 4 & 8 & 16 & 32 & 4 & 8 & 16 & 32 & 4 & 8 & 16 & 32 & 4 & 8 & 16 & 32 & 4 & 8 & 16 & 32 & 4 & 8 & 16 & 32 \\
\bottomrule
\end{tabular}
}
\smallerspacecaption
\smallerspacecaption
\end{table*}

\subsection{Success Against GNN-RE~\cite{GNNRE}}
We tested \myname{} against GNN-RE using a dataset of 37 circuits containing combinations of adders, subtractors, comparators, multipliers, and control logic with bit widths from 
$\{4,8,16,32\}$~\cite{GNNRE_github}. In total, the dataset contains $24$ circuits. We follow the training procedure described in~\cite{GNNRE} to train GNN-RE. Given a circuit, GNN-RE 
classifies the gates in the circuit into one of the five classes (adders, subtractors, comparators, multipliers, and control logic). Hence, our \myname{} tool generates adversarial circuits by perturbing the given circuit 
with the objective of decreasing the classification accuracy of GNN-RE in a black-box setting. We define an adversarial circuit as successful if it results in GNN-RE's classification accuracy of $\leq25\%$.
An important note here is that we consider the more stringent scenario where we generate adversarial examples for the circuits that GNN-RE has seen during training, as opposed to generating adversarial examples for circuits not seen by GNN-RE.

\begin{figure}[!t]
    \centering
    \includegraphics[width=0.48\textwidth]{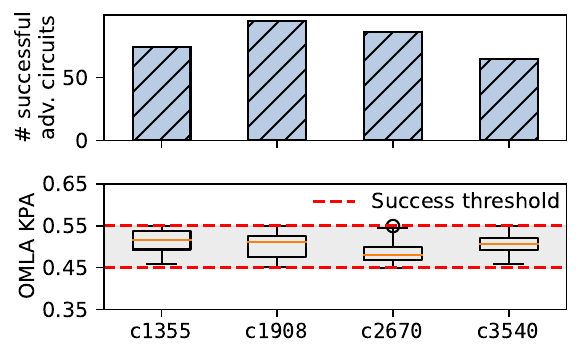}
    \smallerspacecaption
    \smallerspacecaption
    \smallerspacecaption
    \smallerspacecaption
    \caption{Top: Number of successful adversarial circuits against OMLA (higher values: better attack). Bottom: OMLA's accuracy for them (values near $0.5$: better attack).
    }
    \label{fig:OMLA_adv_cnt_and_KPA}
    \smallerspacecaption
    \smallerspacecaption
\end{figure}

Figure~\ref{fig:GNNRE_adv_cnt_and_accuracy} shows the number of successful adversarial examples generated by \myname{} against GNN-RE (top) and GNN-RE's accuracy distribution for those adversarial circuits (bottom). The nomenclature for the circuit labels is explained in Table~\ref{tab:GNNRE_benchmarks}. Even though a single \myname{} RL agent perturbs the circuits seen by GNN-RE during training, it \textbf{successfully fools GNN-RE and results in 
GNN-RE's accuracy to drop to $\bm{<0.25}$, i.e., $\bm{<25\%}$.}

\subsection{Success Against OMLA~\cite{OMLA}}
To assess \myname{} against OMLA, we use the 
the publicly available circuits from the OMLA repository~\cite{OMLA_github}. 
This dataset comprises a sum of $3996$ distinct obfuscated circuits distributed across four specific circuit sets (\texttt{c1355}, \texttt{c1908}, \texttt{c2670}, and \texttt{c3540}) derived from the ISCAS benchmark suite.
We selected these circuits due to OMLA's notable high key prediction accuracy (as high as $95\%$). 
Similar to TrojanSAINT, OMLA also trains separate GNNs for each of the four circuits, so we follow the same training process~\cite{OMLA}. Then, we use a single \myname{} agent to generate adversarial circuits against all four GNNs from OMLA. 
Since OMLA predicts the values of the key bits to unobfuscate the circuit, we use the key prediction accuracy (KPA) to measure the success of OMLA. A KPA of $100\%$ means OMLA has recovered all key bits correctly, and a KPA of $50\%$ means OMLA is no better than a random guess.
So, we define an adversarial circuit as successful if OMLA's KPA is between $50\%$ and $55\%$.

Figure~\ref{fig:OMLA_adv_cnt_and_KPA} shows the number of our successful adversarial circuits against OMLA for all four circuit sets, \texttt{c1355}, \texttt{c1908}, \texttt{c2670}, \texttt{c3540}, (top) and distribution of OMLA's KPA for those adversarial circuits (bottom). 
Both figures clearly illustrate the \textbf{success of \myname{} in perturbing obfuscated circuits that render OMLA no better than a random guess}.

\subsection{Success Against GNN4TJ~\cite{GNN4TJ}}
\begin{figure}[t]
    \centering
    \includegraphics[height=1.3in,trim={0 0.7cm 0 0},clip]{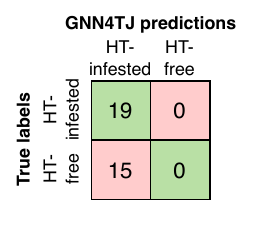}
    \smallerspacecaption
    \caption{Confusion matrix for GNN4TJ predictions.}
    \label{fig:GNN4TJ_conf_matrix}
    \smallerspacecaption
\end{figure}

Here, we analyze the efficacy of GNN4TJ, a GNN-based HT detection technique. Given a circuit, GNN4TJ classifies whether that circuit contains an HT or not. To ensure proper evaluation in our experiments, we used the pre-trained GNN4TJ model as well as the benchmark circuits released by the authors at~\cite{HW2VEC_github}. Additionally, we also included five other HT-free circuits from the OpenCores~\cite{OpenCores} in the evaluation process. Testing a total of $15$ HT-free and $19$ HT-infested circuits, GNN4TJ reported an accuracy of $55.88\%$. 
However, a closer look at the confusion matrix in Figure~\ref{fig:GNN4TJ_conf_matrix} reveals an interesting insight. The false positive rate of GNN4TJ is $\frac{15}{15+0}=100\%$. In other words, GNN4TJ classifies all circuits as HT-infested. Note that of the $15$ circuits that GNN4TJ incorrectly classifies as HT-infested, $9$ are circuits that GNN4TJ has seen during training. This indicates that GNN4TJ has a high bias towards classifying any circuit as HT-infested, making the technique impractical as a HT-detection tool. Since GNN4TJ is highly biased, we do not evaluate it using adversarial examples because no matter what adversarial example (i.e., an HT-infested circuit designed to fool GNN4TJ) is generated, the adversarial example will likely never be classified as HT-free by GNN4TJ making the evaluation moot.

\begin{figure}[t]
    \centering
    \includegraphics[width=0.5\textwidth,trim={0cm 0.2cm 0cm 0.2cm},clip]{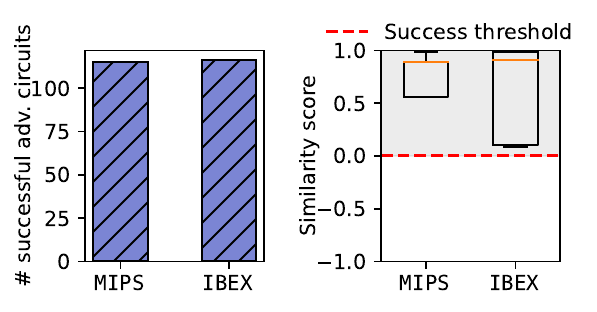}
    \smallerspacecaption
    \smallerspacecaption
    \smallerspacecaption
    \caption{Left: Number of successful adversarial circuits (higher values: better attack). Right:  Distribution of GNN4IP similarity scores for \texttt{MIPS} adversarial circuits with \texttt{IBEX} and vice-versa (higher values: better attack).}
    \label{fig:IP_piracy_case_study_adv_cnt_and_sim_scores_dist}
    \smallerspacecaption
    \smallerspacecaption
\end{figure}

\subsection{Ramifications of Adversarial Examples in Hardware Security}\label{sec:case_studies}
\noindent \textbf{IP Piracy.} So far, we evaluated the efficacy of~\myname{} against GNNs mainly on the benchmarks released for the respective GNNs. Next, we demonstrate the practicality of~\myname{} by showcasing a case study on two large-scale circuits, the Stanford \texttt{MIPS}~\cite{OpenCores_MIPS} and the Google \texttt{IBEX}~\cite{IBEX} processors.\footnote{To ensure compatibility with GNN4IP, we assume full-scan access for these processors.} In particular, we demonstrate the ability of our technique to fool GNN4IP for these large-scale circuits. To that end, we (i) use~\myname{} to generate a variety of perturbed circuits for the \texttt{MIPS} processor and then query GNN4IP to detect piracy between those perturbed circuits and the \texttt{IBEX} processor; and (ii) we generate a variety of perturbed circuits for the \texttt{IBEX} processor and then query GNN4IP to detect piracy between those perturbed circuits and the \texttt{MIPS} processor. We perform this cross-evaluation to determine the false positive rates of GNN4IP on real-world circuits. Figure~\ref{fig:IP_piracy_case_study_adv_cnt_and_sim_scores_dist} shows that~\myname{} generates $>100$ successful adversarial circuits for both the cases in just one hour of training. Even though \texttt{MIPS} and \texttt{IBEX} are completely different circuits,~\myname{} easily fools GNN4IP into classifying them as pirated. Moreover, the distribution of the similarity scores given by GNN4IP to our adversarial circuits of these vastly different processors is also shown in Figure~\ref{fig:IP_piracy_case_study_adv_cnt_and_sim_scores_dist}.\footnote{Note that, unlike Figure~\ref{fig:GNN4IP_sim_scores} in Sec.~\ref{sec:results_GNN4IP}, since the objective of this experiment is to determine the false positive rate of GNN4IP, a successful adversarial circuit is one that fools GNN4IP to classify as pirated, i.e., its similarity score should be higher than $0$.}  It is evident that not only is GNN4IP susceptible to false positives, the magnitude of incorrect classification (as measured by the incorrect high similarity scores) is alarming. This case-study demonstrates the importance of adversarial evaluation in hardware security. Failing to do so can result in a classifier that is highly inaccurate and can lead to circuits being incorrectly flagged as pirated with high confidence.\newline

\noindent \textbf{HT Localization.} Next, we demonstrate the ramifications of adversarial examples for the case of HT localization by showcasing an attack on an \texttt{AES} encryption circuit. \texttt{AES} is a widely-used encryption algorithm (i.e., cipher) and is the first and only publicly accessible cipher approved by the U.S. National Security Agency NSA for top secret information~\cite{AES_NSA}.
For this case study, we design and insert an HT that leaks the secret key when $2^{128}$ encryptions are done.
Then, we use \myname{} to generate adversarial examples for this HT-infested \texttt{AES} and evaluate TrojanSAINT's efficacy in locating the HT. 
\myname{} successfully generates $10$ unique adversarial circuits, each of which results in a $50\%$ or lower score from TrojanSAINT.\footnote{``Score'' here refers to the average of true positive and true negative rates, i.e., the metric used in TrojanSAINT~\cite{TrojanSAINT_ISCAS}.} 
Again, this case study reinforces the need for a thorough evaluation of GNN-based techniques in hardware security, as failing to do so can have disastrous consequences.

\begin{myframe}
\textbf{Results Summary.} All our results validate the efficacy and generality of our technique. 
A single \myname{} RL agent successfully generates adversarial examples against all considered GNNs in just $12$ hours of training.
\end{myframe}

%% file: texfiles/related_work_and_discussion.tex
\section{Related Work and Discussion}\label{sec:related_work}
In this section, we first discuss other works that generate adversarial examples or use reinforcement learning (RL) for security problems and how \myname{} is different from them. Then, we provide a detailed description of another related work that targets the GNNs in hardware security and outline the key points that differentiate our work from it. Finally, we discuss potential countermeasures against \myname{}.
\subsection{Adversarial Examples and RL in Security}\label{sec:related_work_adv_ex_and_RL_in_security}
There has been a plethora of work on attacking systems by generating adversarial examples~\cite{prokos2023squint,wu2023kenku,ahmed2023tubes,chen2023qfa2sr,li2020adversarial,yu2023smack,li2023black,zhang2023capatch,li2019adversarial,eisenhoferno,liu2023x,grosse2017adversarial,ji2021poltergeist,chen2021real,duan2022perception,guo2022specpatch,xu2019exact,chen2017attacking,liu2022evil,zheng2021black,mu2021hard,dai2018adversarial,huang2018adversarial}. Just over the past couple of years, researchers have designed adversarial attacks against perceptual hashing~\cite{prokos2023squint}, automatic speech recognition systems~\cite{wu2023kenku,chen2023qfa2sr,yu2023smack,chen2021real,duan2022perception,guo2022specpatch}, speaker identification~\cite{ahmed2023tubes}, malware detection~\cite{li2023black,huang2018adversarial,grosse2017adversarial,li2020adversarial,li2019adversarial}, image captioning systems~\cite{zhang2023capatch,xu2019exact,chen2017attacking}, image detection~\cite{liu2023x,ji2021poltergeist}, and even an automatic reviewer assignment system used in a top security conference~\cite{eisenhoferno}. All these attacks perturb certain features in the input space to fool a detection/recognition/classification system into producing incorrect outputs. Researchers have also proposed adversarial attacks against GNNs in general~\cite{mu2021hard,dai2018adversarial}, albeit not in hardware security. Since these techniques do not target hardware circuits, they employ existing graph perturbation techniques and work with graphs with a couple thousand nodes. In contrast, our work (i) designs new kinds of perturbations suitable for hardware circuits, (ii) works with circuits that contain up to $258K$ gates (i.e., nodes), and (iii) to the best of our knowledge, is the first work that generates successful adversarial examples against GNN-based classification systems used in hardware security. 

In another direction, researchers have also used RL to devise new attacks and defenses in hardware security~\cite{luo2023autocat,guo2022vulnerability,cui2023macta, gohil2022deterrent,chen2023adatest,gohil2022attrition,sarihi2022hardware,guo2023explorefault,dey2022secure,gohil2023mabfuzz,DETERRENT_TCAD,patnaik2022reinforcement}. However, these works are orthogonal to ours: they target specific problems in hardware security (e.g., HTs, fault injection, cache-timing attacks), whereas our work generates adversarial examples against GNNs used in hardware security. Researchers have also explored the potential of using generative adversarial networks (GANs) to generate adversarial examples~\cite{xiao2018generating,ajorpaz2022evax}. However, such works mainly focus on attacking traditional deep neural networks, so the potential of GANs for generating adversarial examples against GNNs, and specifically GNNs for hardware security, is yet to be explored.

\subsection{Attacking GNNs in Hardware Security}
Different from this work, \texttt{PoisonedGNN}, which targets some of the GNNs in hardware security, assumes a threat model of a backdoor attack, where the attacker has access to the training data, the training process, and the trained model's parameters~\cite{alrahis2023poisonedgnn}. 
This enables it to insert backdoors into the target model, reducing its accuracy during inference. 

Unlike \texttt{PoisonedGNN},~\myname{} is an adversarial example-based attack technique that (i) does not require access to the training procedure, (ii) works under the constraints of not being allowed to change the model parameters or structure, (iii) works with black-box access to the model under attack, and (iv) does not devise specific techniques (i.e., backdoors) dependent on the target GNNs; rather, \myname{} generates adversarial examples in an agnostic manner, using only black-box access to the target GNNs, resulting in thwarting GNNs used in four different hardware security problems.

\vspace{-0.5cm}
\subsection{Potential Countermeasures}\label{sec:countermeasures}
There are several options to protect GNNs against adversarial attacks. (i) Adversarial training involves injecting adversarial examples into the training set such that the trained model can correctly classify the future adversarial examples~\cite{goodfellow2014explaining}. However, research has also shown its limitations in general machine learning settings~\cite{carlini2017adversarial} as well as for GNNs~\cite{gosch2023adversarial,geisler2020reliable}. (ii) Researchers have also devised adversarial perturbation detection techniques as a countermeasure against adversarial attacks~\cite{xu2018characterizing,ioannidis2019graphsac}. However, these techniques are not applicable to \myname{} since our work does not use the typical node/edge perturbation techniques. (iii) Different from the previous heuristic-based approaches, certifiable robustness techniques provide guaranteed defense against adversarial attacks under some assumptions~\cite{bojchevski2019certifiable}. However, such techniques focus on typical node/edge perturbations, different from our perturbations that potentially change the entire graph. More future work is needed to devise such certifiable robustness techniques for our case.

%% file: texfiles/conclusion.tex
\vspace{0.4cm}
\section{Conclusion}\label{sec:conclusion}
Graph neural networks (GNNs) have shown great have shown great potential in addressing several critical hardware security problems. However, we observe that these state-of-the-art GNN-based techniques have lacked thorough evaluation, particularly against the threat of adversarial examples.

Using reinforcement learning (RL), we devised a first-of-its-kind automated technique, \myname{}, that generates adversarial examples against GNNs used in hardware security. To do so, we couldn't rely on existing perturbation-based adversarial example generation methods since working with hardware circuits poses unique constraints (maintaining circuit functionality and obeying circuit design rules) and challenges (scaling to large circuits). We overcame these constraints by developing circuit functionality-preserving perturbations. Moreover, we developed custom optimizations improving the effectiveness and efficiency of our RL agent, allowing it to scale to practical circuits. We also devised a contextual Markov decision process formulation enabling a single RL agent to generate successful adversarial examples against GNNs for four classes of hardware security problems. \myname{} is agnostic to the target GNN architecture and only requires black-box access to the GNNs. 

Experimental results confirm that \myname{}-generated adversarial examples fool all GNNs considered in this work. We also showcase the power of \myname{} in (i) fooling an IP piracy detector for the \texttt{MIPS} and \texttt{IBEX} processors and (ii) creating a circuit compromised with an HT that can leak an \texttt{AES} secret key while evading an HT localization technique.

%% file: texfiles/acknowledgement.tex
\section*{Acknowledgments}
This work was partially supported by the National Science Foundation (NSF CNS--1822848 and NSF DGE--2039610). We thank Dr. Lilas Alrahis from New York University, Abu Dhabi, for helping us set up the OMLA and TrojanSAINT GNNs.

%% file: texfiles/appendix.tex
\section{Appendix}\label{sec:appendix}
\subsection{Pseudocode}\label{sec:pseudocode_and_complexity}

\SetArgSty{textnormal}
\newcommand\mycommfont[1]{\footnotesize\ttfamily\textcolor{codecommentcolor}{#1}}
\SetCommentSty{mycommfont}
\begin{algorithm}[!htb]
\LinesNumbered
\DontPrintSemicolon
\KwInput{List of GNNs and their corresponding circuits}
\KwParam{Number of steps in rollout, $J=32$}
\KwOutput{Trained \myname{} policy $\pi_\theta$}
\While{not converged}{
    $j\leftarrow 0$ \tcp*[l]{rollout step counter}
    $\mathcal{T}\leftarrow\phi$ \tcp*[l]{for storing rollout trajectories}
    \While(\tcp*[h]{\textit{J}-step rollout}){$j<J$}{
        GNN, $T \leftarrow$ Pick a random target GNN\\
        $s_0 \leftarrow c_{\text{GNN}}| \text{random circuit for GNN} $\\
        \For{$t=0,1,2,\ldots, T-1$}{
            $a_{t} \leftarrow \pi_{\theta}(s_t)$\\
            $s_{t+1} \leftarrow P^c(s_{t},a_t)$\\
            $r_t \leftarrow R(s_t,a_t)$\\
            Store trajectories in $\mathcal{T}$\\
            $j\leftarrow j+1$\\
        }
    }
    $\pi_{\theta} \leftarrow$ PPO($\mathcal{T}$) \tcp*[l]{update policy using PPO}
}
\caption{\myname{} pseudocode}
\label{alg:pseudocode}
\end{algorithm}

Algorithm~\ref{alg:pseudocode} details the pseudocode for \myname{}, which runs an iterative process until convergence. In each iteration, (i) the while loop in line 4 rolls out and collects trajectories in $\mathcal{T}$ according to the current policy, $\pi_\theta$, and (ii) the PPO algorithm updates the current policy using the collected trajectories $\mathcal{T}$ (line 13). During the rollout phase, at the beginning of each episode, a random GNN is selected (which also determines the length of that episode, $T$). Then, a random circuit (from the target circuits) is selected for that GNN, which, along with the context of the target GNN, $c_{\text{GNN}}$, constitutes the initial state, $s_0$. Then, the episode is run following the current policy's actions, and the generated trajectory is stored in $\mathcal{T}$, which is then used by the PPO algorithm to update the current policy until convergence. Note that assessing \myname{}'s theoretical computational complexity is non-trivial because it involves RL training and closed-source circuit synthesis algorithms, whose complexities are unknown and difficult to derive. However, the practical runtime required for generating successful adversarial examples against all circuits for all GNNs is less than 12 hours, making \myname{} extremely efficient.

Also note that \myname{} generates successful adversarial examples during the training process itself. This follows the assumptions for adversarial attacks where, for instance, an adversary attempting IP piracy has access to the circuit they wish to pirate as well as to the target GNN they wish to fool so they can query the GNN with the desired circuit as we do during RL training. Moreover, this approach of generating new attacks/defenses during the training process has also been used in recent related works on RL for hardware security, such as~\cite{luo2023autocat,gohil2022attrition}.

\begin{figure}[!tb]
    \centering
    \includegraphics[width=0.5\textwidth,trim={0cm 0.25cm 0cm 0.25cm},clip]{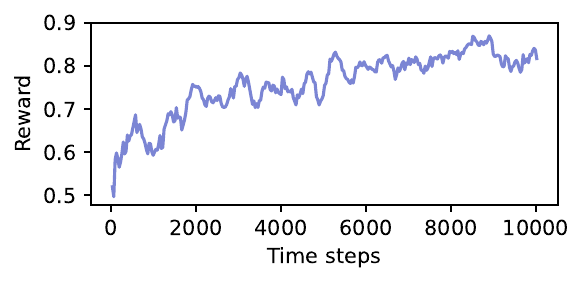}
    \smallerspacecaption
    \smallerspacecaption
    \smallerspacecaption
    \caption{\myname{}'s training reward curve}
    \label{fig:reward_curve}
    \vspace{0.3cm}
\end{figure}

\begin{figure}[!tb]
    \centering
    \includegraphics[width=0.5\textwidth,trim={0cm 0.25cm 0cm 0.2cm},clip]{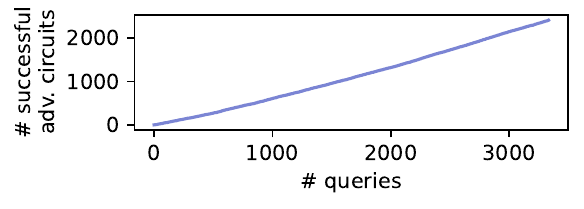}
    \smallerspacecaption
    \smallerspacecaption
    \smallerspacecaption
    \caption{\myname{}'s performance with increasing number of black-box queries}
    \label{fig:num_successful_adv_circs_vs_num_queries}
\end{figure}

\subsection{Convergence and Performance}\label{sec:convergence_and_performance}
Here, we provide results demonstrating the convergence and the performance of \myname{}. Figure~\ref{fig:reward_curve} shows the training reward curve: it is evident that the agent converges to a high reward within $10000$ time steps. Figure~\ref{fig:num_successful_adv_circs_vs_num_queries} illustrates \myname{}'s performance in terms of the number of (non-unique) successful adversarial circuits as a function of the number of black-box queries. As expected, \myname{}'s performance increases with an increasing number of queries. 
\newline

\noindent\textbf{Note.}
This work does not intend to highlight the limitations of any specific technique(s). Instead, it is a reflective endeavor to illustrate how the application of machine learning advancements to address hardware security challenges can potentially introduce new vulnerabilities. Moreover, it emphasizes the importance of comprehensive evaluation and how, as a community, we can approach to mitigate these vulnerabilities.

%% file: main.bbl
\begin{thebibliography}{10}

\bibitem{aiss}
DARPA~Public Affairs.
\newblock {DARPA Selects Teams to Increase Security of Semiconductor Supply Chain}.
\newblock \url{https://www.darpa.mil/news-events/2020-05-27}, 2022.
\newblock [Online; last accessed 17-Oct-2023].

\bibitem{agrawal2007trojan}
Dakshi Agrawal, et~al.
\newblock {Trojan detection using IC fingerprinting}.
\newblock In {\em 2007 IEEE Symposium on Security and Privacy (SP'07)}, pages 296--310. IEEE, 2007.

\bibitem{ahmed2023tubes}
Shimaa Ahmed, et~al.
\newblock {Tubes Among Us: Analog Attack on Automatic Speaker Identification}.
\newblock In {\em 32nd USENIX Security Symposium (USENIX Security 23)}, pages 265--282, 2023.

\bibitem{HW2VEC_github}
{AICPS}.
\newblock {HW2VEC: A Graph Learning Tool for Automating Hardware Security}.
\newblock \url{https://github.com/AICPS/hw2vec}, 2021.
\newblock [Online; last accessed 17-Oct-2023].

\bibitem{ajorpaz2022evax}
Samira~Mirbagher Ajorpaz, et~al.
\newblock Evax: Towards a practical, pro-active \& adaptive architecture for high performance \& security.
\newblock In {\em 2022 55th IEEE/ACM International Symposium on Microarchitecture (MICRO)}, pages 1218--1236. IEEE, 2022.

\bibitem{ABC}
{Alan Mishchenko, et al.}
\newblock {ABC: A System for Sequential Synthesis and Verification}.
\newblock \url{https://people.eecs.berkeley.edu/~alanmi/abc/}, 2007.

\bibitem{alkabani2007active}
Yousra Alkabani et~al.
\newblock {Active Hardware Metering for Intellectual Property Protection and Security}.
\newblock In {\em 16th USENIX Security Symposium (USENIX Security 07)}, volume~20, pages 1--20, 2007.

\bibitem{alrahis2023poisonedgnn}
Lilas Alrahis, et~al.
\newblock {PoisonedGNN: Backdoor attack on graph neural networks-based hardware security systems}.
\newblock {\em IEEE Transactions on Computers}, 2023.

\bibitem{OMLA}
Lilas Alrahis, et~al.
\newblock {OMLA: An oracle-less machine learning-based attack on logic locking}.
\newblock {\em IEEE Transactions on Circuits and Systems II: Express Briefs}, 69(3):1602--1606, 2021.

\bibitem{alrahis2022muxlink}
Lilas Alrahis, et~al.
\newblock {MuxLink: Circumventing learning-resilient mux-locking using graph neural network-based link prediction}.
\newblock In {\em 2022 Design, Automation \& Test in Europe Conference \& Exhibition (DATE)}, pages 694--699. IEEE, 2022.

\bibitem{GNNRE}
Lilas Alrahis, et~al.
\newblock {GNN-RE: Graph neural networks for reverse engineering of gate-level netlists}.
\newblock {\em IEEE Transactions on Computer-Aided Design of Integrated Circuits and Systems}, 41(8):2435--2448, 2021.

\bibitem{becker2013stealthy}
Georg~T Becker, et~al.
\newblock {Stealthy Dopant-Level Hardware Trojans}.
\newblock In {\em International Conference on Cryptographic Hardware and Embedded Systems (CHES)}, pages 197--214. Springer, 2013.

\bibitem{bojchevski2019certifiable}
Aleksandar Bojchevski et~al.
\newblock Certifiable robustness to graph perturbations.
\newblock {\em Advances in Neural Information Processing Systems}, 32, 2019.

\bibitem{cadence_pufs}
{Cadence}.
\newblock {Secret Key Generation with Physically Unclonable Functions}.
\newblock \url{https://community.cadence.com/cadence_blogs_8/b/breakfast-bytes/posts/secret-key-generation-with-physically-unclonable-functions}, 2017.
\newblock [Online; last accessed 14-Feb-2024].

\bibitem{Genus}
{Cadence}.
\newblock {Genus User Guide}, 2019.

\bibitem{cappart2023combinatorial}
Quentin Cappart, et~al.
\newblock {Combinatorial optimization and reasoning with graph neural networks}.
\newblock {\em J. Mach. Learn. Res.}, 24:130--1, 2023.

\bibitem{carlini2017adversarial}
Nicholas Carlini et~al.
\newblock {Adversarial examples are not easily detected: Bypassing ten detection methods}.
\newblock In {\em Proceedings of the 10th ACM workshop on artificial intelligence and security}, pages 3--14, 2017.

\bibitem{chen2021real}
Guangke Chen, et~al.
\newblock {Who is real bob? adversarial attacks on speaker recognition systems}.
\newblock In {\em 2021 IEEE Symposium on Security and Privacy (SP)}, pages 694--711. IEEE, 2021.

\bibitem{chen2023qfa2sr}
Guangke Chen, et~al.
\newblock {QFA2SR: Query-Free Adversarial Transfer Attacks to Speaker Recognition Systems}.
\newblock {\em arXiv preprint arXiv:2305.14097}, 2023.

\bibitem{chen2017attacking}
Hongge Chen, et~al.
\newblock {Attacking visual language grounding with adversarial examples: A case study on neural image captioning}.
\newblock {\em arXiv preprint arXiv:1712.02051}, 2017.

\bibitem{chen2023adatest}
Huili Chen, et~al.
\newblock {AdaTest: Reinforcement learning and adaptive sampling for on-chip hardware Trojan detection}.
\newblock {\em ACM Transactions on Embedded Computing Systems}, 22(2):1--23, 2023.

\bibitem{AES_NSA}
{Committee on National Security Systems}.
\newblock {National Policy on the Use of the Advanced Encryption Standard (AES) to Protect National Security Systems and National Security Information}.
\newblock \url{https://web.archive.org/web/20101106122007/http://csrc.nist.gov/groups/ST/toolkit/documents/aes/CNSS15FS.pdf}, 2003.
\newblock [Online; last accessed 17-Oct-2023].

\bibitem{cui2023macta}
Jiaxun Cui, et~al.
\newblock {MACTA}: A multi-agent reinforcement learning approach for cache timing attacks and detection.
\newblock In {\em The Eleventh International Conference on Learning Representations}, 2023.

\bibitem{dai2018adversarial}
Hanjun Dai, et~al.
\newblock {Adversarial attack on graph structured data}.
\newblock In {\em International conference on machine learning}, pages 1115--1124. PMLR, 2018.

\bibitem{Mircon_IP_piracy}
{Department of Justice}.
\newblock {Attorney General Jeff Sessions Announces New Initiative to Combat Chinese Economic Espionage}.
\newblock \url{https://www.justice.gov/opa/speech/attorney-general-jeff-sessions-announces-new-initiative-combat-chinese-economic-espionage}, 2018.
\newblock [Online; last accessed 17-Oct-2023].

\bibitem{dey2022secure}
Sukanta Dey, et~al.
\newblock Secure physical design.
\newblock {\em Cryptology ePrint Archive}, 2022.

\bibitem{duan2022perception}
Rui Duan, et~al.
\newblock {Perception-Aware Attack: Creating Adversarial Music via Reverse-Engineering Human Perception}.
\newblock In {\em Proceedings of the 2022 ACM SIGSAC Conference on Computer and Communications Security}, pages 905--919, 2022.

\bibitem{eisenhoferno}
Thorsten Eisenhofer, et~al.
\newblock {No more Reviewer\# 2: Subverting Automatic Paper-Reviewer Assignment using Adversarial Learning}.
\newblock 2023.

\bibitem{eykholt2018robust}
Kevin Eykholt, et~al.
\newblock {Robust physical-world attacks on deep learning visual classification}.
\newblock In {\em Proceedings of the IEEE conference on computer vision and pattern recognition}, pages 1625--1634, 2018.

\bibitem{fan2019graph}
Wenqi Fan, et~al.
\newblock {Graph neural networks for social recommendation}.
\newblock In {\em The world wide web conference}, pages 417--426, 2019.

\bibitem{geisler2020reliable}
Simon Geisler, et~al.
\newblock {Reliable graph neural networks via robust aggregation}.
\newblock {\em Advances in Neural Information Processing Systems}, 33:13272--13284, 2020.

\bibitem{GNNRE_github}
{GNNRE}.
\newblock {GNN-RE: Graph Neural Networks for Reverse Engineering of Gate-Level Netlists}.
\newblock \url{https://github.com/DfX-NYUAD/GNN-RE}, 2022.
\newblock [Online; last accessed 17-Oct-2023].

\bibitem{gohil2022attrition}
Vasudev Gohil, et~al.
\newblock {ATTRITION: Attacking Static Hardware Trojan Detection Techniques Using Reinforcement Learning}.
\newblock In {\em Proceedings of the 2022 ACM SIGSAC Conference on Computer and Communications Security}, pages 1275--1289, 2022.

\bibitem{gohil2023mabfuzz}
Vasudev Gohil, et~al.
\newblock {MABFuzz: Multi-Armed Bandit Algorithms for Fuzzing Processors}.
\newblock {\em arXiv preprint arXiv:2311.14594}, 2023.

\bibitem{gohil2022deterrent}
Vasudev Gohil, et~al.
\newblock {DETERRENT: Detecting Trojans using Reinforcement Learning}.
\newblock In {\em Proceedings of the 59th ACM/IEEE Design Automation Conference}, pages 697--702, 2022.

\bibitem{DETERRENT_TCAD}
Vasudev Gohil, et~al.
\newblock {DETERRENT: Detecting Trojans Using Reinforcement Learning}.
\newblock {\em IEEE Transactions on Computer-Aided Design of Integrated Circuits and Systems}, 43(1):57--70, 2024.

\bibitem{goodfellow2014explaining}
Ian~J Goodfellow, et~al.
\newblock Explaining and harnessing adversarial examples.
\newblock {\em arXiv preprint arXiv:1412.6572}, 2014.

\bibitem{gosch2023adversarial}
Lukas Gosch, et~al.
\newblock {Adversarial Training for Graph Neural Networks}.
\newblock {\em arXiv preprint arXiv:2306.15427}, 2023.

\bibitem{grosse2017adversarial}
Kathrin Grosse, et~al.
\newblock {Adversarial examples for malware detection}.
\newblock In {\em Computer Security--ESORICS 2017: 22nd European Symposium on Research in Computer Security, Oslo, Norway, September 11-15, 2017, Proceedings, Part II 22}, pages 62--79. Springer, 2017.

\bibitem{guo2022specpatch}
Hanqing Guo, et~al.
\newblock {Specpatch: Human-in-the-loop adversarial audio spectrogram patch attack on speech recognition}.
\newblock In {\em Proceedings of the 2022 ACM SIGSAC Conference on Computer and Communications Security}, pages 1353--1366, 2022.

\bibitem{guo2023explorefault}
Hao Guo, et~al.
\newblock {ExploreFault: Identifying Exploitable Fault Models in Block Ciphers with Reinforcement Learning}.
\newblock In {\em 2023 60th ACM/IEEE Design Automation Conference (DAC)}, pages 1--6. IEEE, 2023.

\bibitem{guo2022vulnerability}
Hao Guo, et~al.
\newblock {Vulnerability Assessment of Ciphers To Fault Attacks Using Reinforcement Learning}.
\newblock {\em Cryptology ePrint Archive}, 2022.

\bibitem{hicks2010overcoming}
Matthew Hicks, et~al.
\newblock {Overcoming an Untrusted Computing Base: Detecting and Removing Malicious Hardware Automatically}.
\newblock In {\em 2010 IEEE Symposium on Security and Privacy (SP'10)}, pages 159--172. IEEE, 2010.

\bibitem{huang2018adversarial}
Alex Huang, et~al.
\newblock {Adversarial deep learning for robust detection of binary encoded malware}.
\newblock {\em arXiv preprint arXiv:1801.02950}, 2018.

\bibitem{IBEX}
{Ibex}.
\newblock {Ibex RISC-V Core}.
\newblock \url{https://github.com/lowRISC/ibex}, 2023.
\newblock [Online; last accessed 17-Oct-2023].

\bibitem{imeson2013securing}
Frank Imeson, et~al.
\newblock {Securing Computer Hardware Using {3D} Integrated Circuit ({IC}) Technology and Split Manufacturing for Obfuscation}.
\newblock In {\em 22nd USENIX Security Symposium (USENIX Security 13)}, pages 495--510, 2013.

\bibitem{ioannidis2019graphsac}
Vassilis~N Ioannidis, et~al.
\newblock {Graphsac: Detecting anomalies in large-scale graphs}.
\newblock {\em arXiv preprint arXiv:1910.09589}, 2019.

\bibitem{ji2021poltergeist}
Xiaoyu Ji, et~al.
\newblock {Poltergeist: Acoustic adversarial machine learning against cameras and computer vision}.
\newblock In {\em 2021 IEEE Symposium on Security and Privacy (SP)}, pages 160--175. IEEE, 2021.

\bibitem{jin2021adversarial}
Wei Jin, et~al.
\newblock {Adversarial attacks and defenses on graphs}.
\newblock {\em ACM SIGKDD Explorations Newsletter}, 22(2):19--34, 2021.

\bibitem{samsung_228_billion}
{Kharpal, Arjun and Lee, Jhiye}.
\newblock {Samsung to spend \$228 billion on the world’s largest chip facility as part of South Korea tech plan}.
\newblock \url{https://www.cnbc.com/2023/03/15/samsung-to-spend-228-billion-on-the-worlds-largest-chip-facility.html}, 2023.
\newblock [Online; last accessed 17-Oct-2023].

\bibitem{kingma2014adam}
Diederik~P Kingma et~al.
\newblock {Adam: A Method for Stochastic Optimization}.
\newblock {\em arXiv preprint arXiv:1412.6980}, 2014.

\bibitem{TrojanSAINT_ISCAS}
Hazem Lashen, et~al.
\newblock {TrojanSAINT: Gate-Level Netlist Sampling-Based Inductive Learning for Hardware Trojan Detection}.
\newblock In {\em {2023 IEEE International Symposium on Circuits and Systems (ISCAS)}}, pages 1--5, 2023.

\bibitem{li2020adversarial}
Deqiang Li et~al.
\newblock {Adversarial deep ensemble: Evasion attacks and defenses for malware detection}.
\newblock {\em IEEE Transactions on Information Forensics and Security}, 15:3886--3900, 2020.

\bibitem{li2023black}
Heng Li, et~al.
\newblock {Black-box Adversarial Example Attack towards FCG Based Android Malware Detection under Incomplete Feature Information}.
\newblock 2023.

\bibitem{li2019adversarial}
Heng Li, et~al.
\newblock {Adversarial-example attacks toward android malware detection system}.
\newblock {\em IEEE Systems Journal}, 14(1):653--656, 2019.

\bibitem{lillicrap2015continuous}
Timothy~P Lillicrap, et~al.
\newblock {Continuous control with deep reinforcement learning}.
\newblock {\em arXiv preprint arXiv:1509.02971}, 2015.

\bibitem{liu2023x}
Aishan Liu, et~al.
\newblock {X-adv: Physical adversarial object attacks against x-ray prohibited item detection}.
\newblock 2023.

\bibitem{liu2022evil}
Han Liu, et~al.
\newblock {When evil calls: Targeted adversarial voice over ip network}.
\newblock In {\em Proceedings of the 2022 ACM SIGSAC Conference on Computer and Communications Security}, pages 2009--2023, 2022.

\bibitem{luo2023autocat}
Mulong Luo, et~al.
\newblock Autocat: Reinforcement learning for automated exploration of cache-timing attacks.
\newblock In {\em 2023 IEEE International Symposium on High-Performance Computer Architecture (HPCA)}, pages 317--332. IEEE, 2023.

\bibitem{mao2016resource}
Hongzi Mao, et~al.
\newblock {Resource management with deep reinforcement learning}.
\newblock In {\em Proceedings of the 15th ACM workshop on hot topics in networks}, pages 50--56, 2016.

\bibitem{mnih2013playing}
Volodymyr Mnih, et~al.
\newblock {Playing atari with deep reinforcement learning}.
\newblock {\em arXiv preprint arXiv:1312.5602}, 2013.

\bibitem{mu2021hard}
Jiaming Mu, et~al.
\newblock {A hard label black-box adversarial attack against graph neural networks}.
\newblock In {\em Proceedings of the 2021 ACM SIGSAC Conference on Computer and Communications Security}, pages 108--125, 2021.

\bibitem{DoD_supply_chain}
Department of~Defense~(DoD).
\newblock {Securing Defense-Critical Supply Chains}.
\newblock {\em An action plan developed in response to President Biden's Executive Order 14017}, 2022.
\newblock [Online; last accessed 17-Oct-2023].

\bibitem{OMLA_github}
{OMLA}.
\newblock {OMLA: An Oracle-less Machine Learning-based Attack on Logic Locking}.
\newblock \url{https://github.com/DfX-NYUAD/OMLA}, 2021.
\newblock [Online; last accessed 17-Oct-2023].

\bibitem{OpenCores_MIPS}
{OpenCores}.
\newblock {Educational 16-bit MIPS Processor}.
\newblock \url{https://opencores.org/projects/mips_16}, 2013.
\newblock [Online; last accessed 17-Oct-2023].

\bibitem{OpenCores}
{OpenCores}.
\newblock {OpenCores}.
\newblock \url{https://opencores.org/}, 2023.
\newblock [Online; last accessed 17-Oct-2023].

\bibitem{patnaik2022reinforcement}
Satwik Patnaik, et~al.
\newblock {Reinforcement Learning for Hardware Security: Opportunities, Developments, and Challenges}.
\newblock In {\em 2022 19th International SoC Design Conference (ISOCC)}, pages 217--218. IEEE, 2022.

\bibitem{prokos2023squint}
Jonathan Prokos, et~al.
\newblock {Squint Hard Enough: Attacking Perceptual Hashing with Adversarial Machine Learning}.
\newblock In {\em 32nd USENIX Security Symposium (USENIX Security 23)}, pages 211--228, 2023.

\bibitem{rostami2014primer}
Masoud Rostami, et~al.
\newblock {A Primer on Hardware Security: Models, Methods, and Metrics}.
\newblock {\em Proceedings of the IEEE}, 102(8):1283--1295, 2014.

\bibitem{sarihi2022hardware}
Amin Sarihi, et~al.
\newblock Hardware trojan insertion using reinforcement learning.
\newblock In {\em Proceedings of the Great Lakes Symposium on VLSI 2022}, pages 139--142, 2022.

\bibitem{schulman2017proximal}
John Schulman, et~al.
\newblock {Proximal policy optimization algorithms}.
\newblock {\em arXiv preprint arXiv:1707.06347}, 2017.

\bibitem{skorobogatov2012breakthrough}
Sergei Skorobogatov et~al.
\newblock {Breakthrough silicon scanning discovers backdoor in military chip}.
\newblock In {\em International Workshop on Cryptographic Hardware and Embedded Systems (CHES)}, pages 23--40. Springer, 2012.

\bibitem{strokach2020fast}
Alexey Strokach, et~al.
\newblock {Fast and flexible protein design using deep graph neural networks}.
\newblock {\em Cell systems}, 11(4):402--411, 2020.

\bibitem{synopsys_pufs}
{Synopsys}.
\newblock {Physically Unclonable Function (PUF) Solution for ARC EM Processors}.
\newblock \url{https://www.synopsys.com/dw/doc.php/ds/cc/intrinsic-ID_PUF_ARC_EM.pdf}.
\newblock [Online; last accessed 14-Feb-2024].

\bibitem{Design_Compiler}
{Synopsys}.
\newblock {Design Compiler User Guide}.
\newblock Version O-2018.06-SP3, 2018.

\bibitem{trippel2021bomberman}
Timothy Trippel, et~al.
\newblock {Bomberman: Defining and Defeating Hardware Ticking Timebombs at Design-time}.
\newblock In {\em 2021 IEEE Symposium on Security and Privacy (SP'21)}, pages 970--986. IEEE, 2021.

\bibitem{TrojanSAINT_github}
{TrojanSAINT}.
\newblock {TrojanSAINT: Gate-Level Netlist Sampling-BasedInductive Learning for Hardware Trojan Detection}.
\newblock \url{https://github.com/DfX-NYUAD/TrojanSAINT}, 2023.
\newblock [Online; last accessed 17-Oct-2023].

\bibitem{TrustHub}
{Trust-Hub}.
\newblock {Trust-Hub}.
\newblock \url{https://www.trust-hub.org/}, 2022.
\newblock [Online; last accessed 17-Oct-2023].

\bibitem{GAT}
Petar Veli{\v{c}}kovi{\'c}, et~al.
\newblock {Graph attention networks}.
\newblock {\em arXiv preprint arXiv:1710.10903}, 2017.

\bibitem{companies_pufs}
{Verdict}.
\newblock {Cybersecurity: who are the leaders in physically unclonable functions (PUFs) for the technology industry?}
\newblock \url{https://www.verdict.co.uk/innovators-cybersecurity-physically-unclonable-functions-pufs-technology/}, 2023.
\newblock [Online; last accessed 14-Feb-2024].

\bibitem{waksman2013fanci}
Adam Waksman, et~al.
\newblock {FANCI: Identification of Stealthy Malicious Logic Using Boolean Functional Analysis}.
\newblock In {\em Proceedings of the 2013 ACM SIGSAC Conference on Computer and Communications Security}, pages 697--708, 2013.

\bibitem{wu2023kenku}
Xinghui Wu, et~al.
\newblock {$\{$KENKU$\}$: Towards Efficient and Stealthy Black-box Adversarial Attacks against $\{$ASR$\}$ Systems}.
\newblock In {\em 32nd USENIX Security Symposium (USENIX Security 23)}, pages 247--264, 2023.

\bibitem{xiao2018generating}
Chaowei Xiao, et~al.
\newblock Generating adversarial examples with adversarial networks.
\newblock {\em arXiv preprint arXiv:1801.02610}, 2018.

\bibitem{GIN}
Keyulu Xu, et~al.
\newblock {How powerful are graph neural networks?}
\newblock {\em arXiv preprint arXiv:1810.00826}, 2018.

\bibitem{xu2018characterizing}
Xiaojun Xu, et~al.
\newblock {Characterizing malicious edges targeting on graph neural networks}.
\newblock 2018.

\bibitem{xu2019exact}
Yan Xu, et~al.
\newblock {Exact adversarial attack to image captioning via structured output learning with latent variables}.
\newblock In {\em Proceedings of the IEEE/CVF Conference on Computer Vision and Pattern Recognition}, pages 4135--4144, 2019.

\bibitem{yang2016a2}
Kaiyuan Yang, et~al.
\newblock {A2: Analog Malicious Hardware}.
\newblock In {\em 2016 IEEE Symposium on Security and Privacy (SP'16)}, pages 18--37. IEEE, 2016.

\bibitem{GNN4TJ}
Rozhin Yasaei, et~al.
\newblock {GNN4TJ: Graph Neural Networks for Hardware Trojan Detection at Register Transfer Level}.
\newblock In {\em 2021 Design, Automation \& Test in Europe Conference \& Exhibition (DATE)}, pages 1504--1509, 2021.

\bibitem{GNN4IP}
Rozhin Yasaei, et~al.
\newblock {GNN4IP: Graph Neural Network for Hardware Intellectual Property Piracy Detection}.
\newblock In {\em 2021 58th ACM/IEEE Design Automation Conference (DAC)}, pages 217--222, 2021.

\bibitem{yu2023smack}
Zhiyuan Yu, et~al.
\newblock {$\{$SMACK$\}$: Semantically Meaningful Adversarial Audio Attack}.
\newblock In {\em 32nd USENIX Security Symposium (USENIX Security 23)}, pages 3799--3816, 2023.

\bibitem{GraphSAINT}
Hanqing Zeng, et~al.
\newblock {Graphsaint: Graph sampling based inductive learning method}.
\newblock {\em arXiv preprint arXiv:1907.04931}, 2019.

\bibitem{zhang2023capatch}
Shibo Zhang, et~al.
\newblock {$\{$CAPatch$\}$: Physical Adversarial Patch against Image Captioning Systems}.
\newblock In {\em 32nd USENIX Security Symposium (USENIX Security 23)}, pages 679--696, 2023.

\bibitem{zheng2021black}
Baolin Zheng, et~al.
\newblock {Black-box adversarial attacks on commercial speech platforms with minimal information}.
\newblock In {\em Proceedings of the 2021 ACM SIGSAC Conference on Computer and Communications Security}, pages 86--107, 2021.

\end{thebibliography}
